\documentclass[MAL,biber]{nowfnt} %

\usepackage{mwe}

\usepackage{longtable}
\usepackage{todonotes}
\usepackage{subcaption}

\usepackage{amsthm}
\usepackage{amsmath}
\usepackage{amsfonts}
\usepackage{amssymb}
\usepackage{comment}
\usepackage[framemethod=tikz]{mdframed}

\usepackage{appendix}
\usepackage{chngcntr}
\usepackage{etoolbox}

\usepackage[ruled]{algorithm2e}
\usepackage{algpseudocode}

\usepackage{lettrine}

\usepackage{mathtools}%
\usepackage{amsmath}

\newcommand{\bb}[1]{\mathbf{#1}}
\newcommand{\bbb}{\bb{b}}
\newcommand{\ba}{\bb{a}}
\newcommand{\bx}{\bb{x}}

\newcommand{\by}{\bb{y}}

\newcommand{\bw}{\bb{w}}

\newcommand{\bg}{\bb{g}}
\newcommand{\bh}{\bb{h}}

\newcommand{\bu}{\bb{u}}
\newcommand{\bp}{\bb{p}}

\newcommand{\bz}{\bb{z}}

\newcommand{\bT}{\boldsymbol{\theta}}

\newcommand{\boldeta}{\boldsymbol{\eta}}

\newcommand{\bphi}{\boldsymbol{\phi}}
\newcommand{\beps}{\boldsymbol{\epsilon}}
\newcommand{\bepsilon}{\boldsymbol{\epsilon}}

\newcommand{\bsigma}{\boldsymbol{\sigma}}
\newcommand{\bmu}{\boldsymbol{\mu}}

\newcommand{\bs}{\mathbf{s}}
\newcommand{\bmm}{\mathbf{m}}

\newcommand{\bL}{\bb{L}}

\newcommand{\bW}{\bb{W}}

\newcommand{\bI}{\bb{I}}

\newcommand{\bX}{\bb{X}}
\newcommand{\bY}{\bb{Y}}
\newcommand{\bZ}{\bb{Z}}

\newcommand{\pT}{p_{\bT}}

\newcommand{\qPhi}{q_{\bphi}}
\newcommand{\qP}{q_{\bphi}}
\newcommand{\qD}{q_\mathcal{D}}
\newcommand{\qDP}{q_{\mathcal{D},\bphi}}

\newcommand{\NeuralNet}{\text{NeuralNet}}
\newcommand{\EncoderNeuralNet}{\text{EncoderNeuralNet}_{\bphi}}
\newcommand{\DecoderNeuralNet}{\text{DecoderNeuralNet}_{\bT}}

\newcommand{\dP}{d_{\bphi}}

\newcommand{\Exp}[2]{\mathbb{E}_{#1}\left[#2\right]}

\newcommand{\ELBO}{\mathcal{L}_{\bT,\bphi}}
\newcommand{\TELBO}{\tilde{\mathcal{L}}_{\bT,\bphi}}

\DeclareMathOperator*{\argmin}{argmin}
\DeclareMathOperator*{\argmax}{argmax}

\title{An Introduction to\\
Variational Autoencoders}

\maintitleauthorlist{
Diederik P. Kingma \\
Google\\
durk@google.com\\
\and
Max Welling \\
Universiteit van Amsterdam, Qualcomm \\
mwelling@qti.qualcomm.com
}

\issuesetup
{%
 copyrightowner={A.~Heezemans and M.~Casey},
 volume        = xx,
 issue         = xx,
 pubyear       = 2019,
 isbn          = xxx-x-xxxxx-xxx-x,
 eisbn         = xxx-x-xxxxx-xxx-x,
 doi           = 10.1561/XXXXXXXXX,
 firstpage     = 1, %
 lastpage      = 18
 }

\author[1]{Kingma, Diederik P.}
\author[2,3]{Welling, Max}

\affil[1]{Google; durk@google.com}
\affil[2]{Universiteit van Amsterdam}
\affil[3]{Qualcomm; mwelling@qti.qualcomm.com}

\articledatabox{\nowfntstandardcitation}
\begin{document}

\makeabstracttitle

\begin{abstract}
Variational autoencoders provide a principled framework for learning deep latent-variable models and corresponding inference models. In this work, we provide an introduction to variational autoencoders and some important extensions.
\end{abstract}

\chapter{Introduction}
\label{chap:introduction}

\section{Motivation}

One major division in machine learning is generative versus discriminative modeling. While in discriminative modeling one aims to learn a predictor given the observations, in generative modeling one aims to solve the more general problem of learning a joint distribution over all the variables. A generative model simulates how the data is generated in the real world. ``Modeling'' is understood in almost every science as unveiling this generating process by hypothesizing theories and testing these theories through observations. For instance, when meteorologists model the weather they use highly complex partial differential equations to express the underlying physics of the weather. Or when an astronomer models the formation of galaxies s/he encodes in his/her equations of motion the physical laws under which stellar bodies interact. The same is true for biologists, chemists, economists and so on. Modeling in the sciences is in fact almost always generative modeling.

There are many reasons why generative modeling is attractive. First, we can express physical laws and constraints into the generative process while details that we don't know or care about, i.e. nuisance variables, are treated as noise. The resulting models are usually highly intuitive and  interpretable and by testing them against observations we can confirm or reject our theories about how the world works. 

Another reason for trying to understand the generative process of data is that it naturally expresses causal relations of the world. Causal relations have the great advantage that they generalize much better to new situations than mere correlations. For instance, once we understand the generative process of an earthquake, we can use that knowledge both in California and in Chile. 

To turn a generative model into a discriminator, we need to use Bayes rule. For instance, we have a generative model for an earthquake of type A and another for type B, then seeing which of the two describes the data best we can compute a probability for whether earthquake A or B happened. Applying Bayes rule is however often computationally expensive. 

\looseness=-1In discriminative methods we directly learn a map in the same direction as we intend to make future predictions in. This is in the opposite direction than the generative model. For instance, one can argue that an image is generated in the world by first identifying the object, then generating the object in 3D and then projecting it onto an pixel grid. A discriminative model takes these pixel values directly as input and maps them to the labels. While generative models can learn efficiently from data, they also tend to make stronger assumptions on the data than their purely discriminative counterparts, often leading to higher asymptotic bias~\citep{banerjee2007analysis} when the model is wrong. For this reason, if the model is wrong (and it almost always is to some degree!), if one is solely interested in learning to discriminate, and one is in a regime with a sufficiently large amount of data, then purely discriminative models typically will lead to fewer errors in discriminative tasks. Nevertheless, depending on how much data is around, it may pay off to study the data generating process as a way to guide the training of the discriminator, such as a classifier. For instance, one may have few labeled examples and many more unlabeled examples. In this semi-supervised learning setting, one can use the generative model of the data to improve  classification~\citep{kingma2014semi,sonderby2016train}.

\looseness=-1Generative modeling can be useful more generally. One can think of it as an auxiliary task. For instance, predicting the immediate future may help us build useful  abstractions of the world that can be used for multiple prediction tasks downstream. This quest for disentangled, semantically meaningful, statistically independent and causal factors of variation in data is generally known as unsupervised representation learning, and the variational autoencoder (VAE) has been extensively employed for that purpose. Alternatively, one may view this as an implicit form of regularization: by forcing the representations to be meaningful for data generation, we bias the inverse of that process, which maps from input to representation, into a certain mould. The auxiliary task of predicting the world is used to better understand the world at an abstract level and thus to better make downstream predictions. 

The VAE can be viewed as two coupled, but independently parameterized models: the encoder or recognition model, and the decoder or generative model. These two models support each other. The recognition model delivers to the generative model an approximation to its posterior over latent random variables, which it needs to update its parameters inside an iteration of ``expectation maximization'' learning. Reversely, the generative model is a scaffolding of sorts for the recognition model to learn meaningful representations of the data, including possibly class-labels. The recognition model is the approximate inverse of the generative model according to Bayes rule. 

One advantage of the VAE framework, relative to ordinary Variational Inference (VI), is that the recognition model (also called inference model) is now a (stochastic) function of the input variables. This in contrast to VI where each data-case has a separate variational distribution, which is inefficient for large data-sets. The recognition model uses one set of parameters to model the relation between input and latent variables and as such is called ``amortized inference''. This recognition model can be arbitrary complex but is still reasonably fast because by construction it can be done using a single feedforward pass from input to latent variables. However the price we pay is that this sampling induces sampling noise in the gradients required for learning. Perhaps the greatest contribution of the VAE framework is the realization that we can counteract this variance by using what is now known as the ``reparameterization trick'', a simple procedure to reorganize our gradient computation that reduces variance in the gradients.

The VAE is inspired by the Helmholtz Machine \citep{dayan1995helmholtz} which was perhaps the first model that employed a recognition model. However, its wake-sleep algorithm was inefficient and didn't optimize a single objective. The VAE learning rules instead follow from a single approximation to the maximum likelihood objective.   

VAEs marry graphical models and deep learning. The generative model is a Bayesian network of the form $p(\bx|\bz) p(\bz)$, or, if there are multiple stochastic latent layers, a hierarchy such as $p(\bx|\bz_L) p(\bz_L|\bz_{L-1})$ $... p(\bz_1|\bz_0)$. Similarly, the recognition model is also a  conditional Bayesian network of the form $q(\bz|\bx)$ or as a hierarchy, such as $q(\bz_0|\bz_1) ... q(\bz_L|X)$. But inside each conditional may hide a complex (deep) neural network, e.g. $\bz|\bx \sim f(\bx,\beps)$, with $f$ a neural network mapping and $\beps$ a noise random variable. Its learning algorithm is a mix of classical (amortized, variational) expectation maximization but through the reparameterization trick ends up backpropagating through the many layers of the deep neural networks embedded inside of it. 

Since its inception, the VAE framework has been extended in many directions, e.g. to dynamical models~\citep{johnson2016composing}, models with attention~\citep{gregor2015draw}, models with multiple levels of stochastic latent variables \citep{kingma2016improving}, and many more. It has proven itself as a fertile framework to build new models in. More recently, another generative modeling paradigm has gained significant attention: the generative adversarial network (GAN)~\citep{goodfellow2014generative}. VAEs and GANs seem to have complementary properties: while GANs can generate images of high subjective perceptual quality, they tend to lack full support over the data~\citep{grover2018flow}, as opposed to likelihood-based generative models. VAEs, like other likelihood-based models, generate more dispersed samples, but are better density models in terms of the likelihood criterion. As such many hybrid models have been proposed to try to represent the best of both worlds \citep{dumoulin2016adversarially,grover2018flow,rosca2018distribution}.

As a community we seem to have embraced the fact that generative models and unsupervised learning play an important role in building intelligent machines. We hope that the VAE provides a useful piece of that puzzle.

\section{Aim}

The framework of \emph{variational autoencoders} (VAEs)~\citep{kingma2013auto,rezende2014stochastic} provides a principled method for jointly learning \emph{deep latent-variable models} and corresponding inference models using stochastic gradient descent. The framework has a wide array of applications from generative modeling, semi-supervised learning to representation learning. 

This work is meant as an expanded version of our earlier work \citep{kingma2013auto}, allowing us to explain the topic in finer detail and to discuss a selection of important follow-up work. This is \emph{not} aimed to be a comprehensive review of all related work. We assume that the reader has basic knowledge of algebra, calculus and probability theory.

In this chapter we discuss background material: probabilistic models, directed graphical models, the marriage of directed graphical models with neural networks, learning in fully observed models and deep latent-variable models (DLVMs). In chapter~\ref{chap:vaes} we explain the basics of VAEs. In chapter \ref{chap:advanced_q} we explain advanced inference techniques, followed by an explanation of advanced generative models in chapter~\ref{chap:advanced_p}.  Please refer to section~\ref{sec:notdef} for more information on mathematical notation.

\section{Probabilistic Models and Variational Inference}

In the field of machine learning, we are often interested in learning probabilistic models of various natural and artificial phenomena from data. Probabilistic models are mathematical descriptions of such phenomena. They are useful for understanding such phenomena, for prediction of unknowns in the future, and for various forms of assisted or automated decision making. As such, probabilistic models formalize the notion of knowledge and skill, and are central constructs in the field of machine learning and AI.

As probabilistic models contain unknowns and the data rarely paints a complete picture of the unknowns, we typically need to assume some level of uncertainty over aspects of the model. The degree and nature of this uncertainty is specified in terms of (conditional) probability distributions. 
Models may consist of both continuous-valued variables and discrete-valued variables. The, in some sense, most complete forms of probabilistic models specify all correlations and higher-order dependencies between the variables in the model, in the form of a joint probability distribution over those variables.

Let's use $\bx$ as the vector representing the set of all observed variables whose joint distribution we would like to model. Note that for notational simplicity and to avoid clutter, we use lower case bold (e.g. $\bx$) to denote the underlying set of observed random variables, i.e. flattened and concatenated such that the set is represented as a single vector. See section~\ref{sec:notdef} for more on notation.

We assume the observed variable $\bx$ is a random sample from an \emph{unknown underlying process}, whose true (probability) distribution $p^*(\bx)$ is unknown. We attempt to approximate this underlying process with a chosen model $\pT(\bx)$, with parameters $\bT$:
\begin{align}
\bx &\sim \pT(\bx)
\end{align}
\emph{Learning} is, most commonly, the process of searching for a value of the parameters $\bT$ such that the probability distribution function given by the model, $\pT(\bx)$, approximates the true distribution of the data, denoted by $p^*(\bx)$, such that for any observed $\bx$:
\begin{align}
\pT(\bx) \approx p^*(\bx)
\end{align}

Naturally, we wish $\pT(\bx)$ to be sufficiently \emph{flexible} to be able to adapt to the data, such that we have a chance of obtaining a sufficiently accurate model. At the same time, we wish to be able to incorporate knowledge about the distribution of data into the model that is known a priori.

\subsection{Conditional Models}

Often, such as in case of classification or regression problems, we are not interested in learning an unconditional model $\pT(\bx)$, but a conditional model $\pT(\by|\bx)$ that approximates the underlying conditional distribution $p^*(\by|\bx)$: a distribution over the values of variable $\by$, conditioned on the value of an observed variable $\bx$. In this case, $\bx$ is often called the \emph{input} of the model.
Like in the unconditional case, a model $\pT(\by|\bx)$ is chosen, and optimized to be close to the unknown underlying distribution, such that for any $\bx$ and $\by$:
\begin{align}
\pT(\by|\bx) \approx p^*(\by|\bx)
\end{align}
A relatively common and simple example of conditional modeling is image classification, where $\bx$ is an image, and $\by$ is the image's class, as labeled by a human, which we wish to predict. In this case, $\pT(\by|\bx)$ is typically chosen to be a categorical distribution, whose parameters are computed from $\bx$.

Conditional models become more difficult to learn when the predicted variables are very high-dimensional, such as images, video or sound. One example is the reverse of the image classification problem: prediction of a distribution over images, conditioned on the class label. Another example with both high-dimensional input, and high-dimensional output, is time series prediction, such as text or video prediction.

To avoid notational clutter we will often assume unconditional modeling, but one should always keep in mind that the methods introduced in this work are, in almost all cases, equally applicable to conditional models. The data on which the model is conditioned, can be treated as inputs to the model, similar to the parameters of the model, with the obvious difference that one doesn't optimize over their value.

\section{Parameterizing Conditional Distributions with Neural Networks}
Differentiable feed-forward neural networks, from here just called \emph{neural networks}, are a particularly flexible and computationally scalable type of function approximator. Learning of models based on neural networks with multiple 'hidden' layers of artificial neurons is often referred to as \emph{deep learning} ~\citep{goodfellow2016deeplearning,lecun2015deep}. A particularly interesting application is probabilistic models, i.e. the use of neural networks for probability density functions (PDFs) or probability mass functions (PMFs) in probabilistic models. Probabilistic models based on neural networks are computationally scalable since they allow for stochastic gradient-based optimization which, as we will explain, allows scaling to large models and large datasets. We will denote a deep neural network as a vector function: $\text{NeuralNet}(.)$.

At the time of writing, deep learning has been shown to work well for a large variety of classification and regression problems, as summarized in~\citep{lecun2015deep,goodfellow2016deeplearning}. 
In case of neural-network based image classification~\cite{lecun1998gradient}, for example, neural networks parameterize a categorical distribution $\pT(y|\bx)$ over a class label $y$, conditioned on an image $\bx$.
\begin{align}
\bp &= \text{NeuralNet}(\bx)\\
\pT(y|\bx) &= \text{Categorical}(y; \bp)
\end{align}
where the last operation of $\text{NeuralNet}(.)$ is typically a softmax() function such that $\sum_i p_i = 1$. 

\section{Directed Graphical Models and Neural Networks}
\label{sec:directedgraphicalmodel}
We work with \emph{directed} probabilistic models, also called directed \emph{probabilistic graphical models} (PGMs), or \emph{Bayesian networks}. Directed graphical models are a type of probabilistic models where all the variables are topologically organized into a directed acyclic graph. The joint distribution over the variables of such models factorizes as a product of prior and conditional distributions: 
\begin{align}
\pT(\bx_1,...,\bx_M) = \prod_{j=1}^M \pT(\bx_j | Pa(\bx_j)) 
\label{eq:dgm}
\end{align}
where $Pa(\bx_j)$ is the set of parent variables of node $j$ in the directed graph. For non-root-nodes, we condition on the parents. For root nodes, the set of parents is the empty set, such that the distribution is unconditional.

Traditionally, each conditional probability distribution $\pT(\bx_j | Pa(\bx_j))$ is parameterized as a lookup table or a linear model \citep{koller2009probabilistic}. As we explained above, a more flexible way to parameterize such conditional distributions is with neural networks. In this case, neural networks take as input the parents of a variable in a directed graph, and produce the distributional parameters $\boldeta$ over that variable:
\begin{align}
\boldeta &= \text{NeuralNet}(Pa(\bx))\\
\pT(\bx|Pa(\bx)) &= \pT(\bx|\boldeta)
\label{eq:nnconditoinal}
\end{align}

We will now discuss how to learn the parameters of such models, if all the variables are observed in the data.

\section{Learning in Fully Observed Models with Neural Nets}

If all variables in the directed graphical model are observed in the data, then we can compute and differentiate the log-probability of the data under the model, leading to relatively straightforward optimization.

\subsection{Dataset}
We often collect a dataset $\mathcal{D}$ consisting of $N \geq 1$ datapoints:
\begin{align}
\mathcal{D}  = \{\bx^{(1)}, \bx^{(2)}, ..., \bx^{(N)} \} \equiv \{\bx^{(i)}\}_{i=1}^N \equiv \bx^{(1:N)}
\end{align}
The datapoints are assumed to be independent samples from an unchanging underlying distribution. In other words, the dataset is assumed to consist of distinct, independent measurements from the same (unchanging) system. In this case, the observations $\mathcal{D} = \{\bx^{(i)}\}_{i=1}^N$ are said to be \emph{i.i.d.}, for \emph{independently and identically distributed}. Under the i.i.d. assumption, the probability of the datapoints given the parameters factorizes as a product of individual datapoint probabilities. The log-probability assigned to the data by the model is therefore given by:
\begin{align}
\log \pT(\mathcal{D}) 
&= \sum_{\bx \in \mathcal{D}} \log \pT(\bx)
\label{eq:iid2}
\end{align}

\subsection{Maximum Likelihood and Minibatch SGD}
The most common criterion for probabilistic models is \emph{maximum log-likelihood} (ML). As we will explain, maximization of the log-likelihood criterion is equivalent to minimization of a Kullback Leibler divergence between the data and model distributions.

Under the ML criterion, we attempt to find the parameters $\bT$ that maximize the sum, or equivalently the average, of the log-probabilities assigned to the data by the model. With i.i.d. dataset $\mathcal{D}$ of size $N_\mathcal{D}$, the maximum likelihood objective is to maximize the log-probability given by equation \eqref{eq:iid2}.

Using calculus' chain rule and automatic differentiation tools, we can efficiently compute gradients of this objective, i.e. the first derivatives of the objective w.r.t. its parameters $\bT$. We can use such gradients to iteratively hill-climb to a local optimum of the ML objective.
If we compute such gradients using all datapoints, $\nabla_{\bT} \log \pT(\mathcal{D})$, then this is known as \emph{batch} gradient descent. Computation of this derivative is, however, an expensive operation for large dataset size $N_\mathcal{D}$, since it scales linearly with $N_\mathcal{D}$.

A more efficient method for optimization is \emph{stochastic gradient descent} (SGD) (section~\ref{sec:sgd}), which uses randomly drawn minibatches of data $\mathcal{M} \subset \mathcal{D}$ of size $N_\mathcal{M}$. With such minibatches we can form an unbiased estimator of the ML criterion:
\begin{align}
\frac{1}{N_\mathcal{D}} \log \pT(\mathcal{D})
\simeq \frac{1}{N_\mathcal{M}} \log \pT(\mathcal{M}) 
= \frac{1}{N_\mathcal{M}} \sum_{\bx \in \mathcal{M}} \log \pT(\bx)
\end{align}
The $\simeq$ symbol means that one of the two sides is an \emph{unbiased estimator} of the other side. So one side (in this case the right-hand side) is a random variable due to some noise source, and the two sides are equal when averaged over the noise distribution. The noise source, in this case, is the randomly drawn minibatch of data $\mathcal{M}$.  The unbiased estimator $\log \pT(\mathcal{M})$ is differentiable, yielding the unbiased stochastic gradients:
\begin{align}
\frac{1}{N_\mathcal{D}} \nabla_{\bT} \log \pT(\mathcal{D})
\simeq \frac{1}{N_\mathcal{M}} \nabla_{\bT} \log \pT(\mathcal{M})
= \frac{1}{N_\mathcal{M}} \sum_{\bx \in \mathcal{M}} \nabla_{\bT} \log \pT(\bx)
\end{align}
These gradients can be plugged into stochastic gradient-based optimizers; see section~\ref{sec:sgd} for further discussion. In a nutshell, we can optimize the objective function by repeatedly taking small steps in the direction of the stochastic gradient.

\subsection{Bayesian inference}

From a Bayesian perspective, we can improve upon ML through \emph{maximum a posteriori} (MAP) estimation (section section ~\ref{sec:map}), or, going even further, inference of a full approximate posterior distribution over the parameters (see section ~\ref{sec:bayesianinference}). 

\section{Learning and Inference in Deep Latent Variable Models}

\subsection{Latent Variables}
\label{sec:latentvariables}

We can extend fully-observed directed models, discussed in the previous section, into directed models with \emph{latent variables}. Latent variables are variables that are part of the model, but which we don't observe, and are therefore not part of the dataset. We typically use $\bz$ to denote such latent variables. In case of unconditional modeling of observed variable $\bx$, the directed graphical model would then represent a joint distribution $\pT(\bx, \bz)$ over both the observed variables $\bx$ and the latent variables $\bz$. The marginal distribution over the observed variables $\pT(\bx)$, is given by:
\begin{align}
\pT(\bx) = \int \pT(\bx,\bz) \,d\bz
\label{eq:marginal}
\end{align}
This is also called the (single datapoint) \emph{marginal likelihood} or the \emph{model evidence}, when taken as a function of $\bT$.

Such an implicit distribution over $\bx$ can be quite flexible. If $\bz$ is discrete and $\pT(\bx|\bz)$ is a Gaussian distribution, then $\pT(\bx)$ is a mixture-of-Gaussians distribution. For continuous $\bz$, $\pT(\bx)$ can be seen as an infinite mixture, which are potentially more powerful than discrete mixtures. Such marginal distributions are also called compound probability distributions.

\subsection{Deep Latent Variable Models}
\label{sec:dlvms}

We use the term \emph{deep latent variable model} (DLVM) to denote a latent variable model $\pT(\bx,\bz)$ whose distributions are parameterized by neural networks. Such a model can be conditioned on some context, like $\pT(\bx,\bz|\by)$. One important advantage of DLVMs, is that even when each factor (prior or conditional distribution) in the directed model is relatively simple (such as conditional Gaussian), the marginal distribution $\pT(\bx)$ can be very complex, i.e. contain almost arbitrary dependencies. This expressivity makes deep latent-variable models attractive for approximating complicated underlying distributions $p^*(\bx)$.

Perhaps the simplest, and most common, DLVM is one that is specified as factorization with the following structure:
\begin{align}
\pT(\bx,\bz) = \pT(\bz) \pT(\bx|\bz)
\end{align}
where $\pT(\bz)$ and/or $\pT(\bx|\bz)$ are specified. The distribution $p(\bz)$ is often called the \emph{prior distribution} over $\bz$, since it is not conditioned on any observations.

\subsection{Example DLVM for multivariate Bernoulli data}

A simple example DLVM, used in ~\citep{kingma2013auto} for binary data $\bx$, is with a spherical Gaussian latent space, and a factorized Bernoulli observation model:
\begin{align}
p(\bz) &= \mathcal{N}(\bz; 0,\bI)\\
\mathbf{p} &= \DecoderNeuralNet(\bz)\\
\log p(\bx|\bz) &=  \sum_{j=1}^D \log p(x_j|\bz) = \sum_{j=1}^D \log \text{Bernoulli}(x_j; p_j)\\
&= \sum_{j=1}^D x_j \log p_j + (1-x_j) \log (1-p_j)
\end{align}
where $\forall p_j \in \bp: 0 \leq p_j \leq 1$ (e.g. implemented through a sigmoid nonlinearity as the last layer of the $\DecoderNeuralNet(.)$), where $D$ is the dimensionality of $\bx$, and $\text{Bernoulli}(.;p)$ is the probability mass function (PMF) of the Bernoulli distribution.

\section{Intractabilities}

The main difficulty of maximum likelihood learning in DLVMs is that the marginal probability of data under the model is typically intractable. This is due to the integral in equation~\eqref{eq:marginal} for computing the marginal likelihood (or model evidence), $\pT(\bx) = \int \pT(\bx,\bz) \,d\bz$, not having an analytic solution or efficient estimator. Due to this intractability, we cannot differentiate it w.r.t. its parameters and optimize it, as we can with fully observed models.

The intractability of $\pT(\bx)$, is related to the intractability of the posterior distribution $\pT(\bz|\bx)$. Note that the joint distribution $\pT(\bx,\bz)$ is efficient to compute, and that the densities are related through the basic identity:
\begin{align}
\pT(\bz|\bx) = \frac{\pT(\bx,\bz)}{\pT(\bx)}
\end{align}
Since $\pT(\bx,\bz)$ \emph{is} tractable to compute, a tractable marginal likelihood $\pT(\bx)$ leads to a tractable posterior $\pT(\bz|\bx)$, and vice versa. Both are intractable in DLVMs.

Approximate inference techniques (see also section ~\ref{sec:altneratives}) allow us to approximate the posterior $\pT(\bz|\bx)$ and the marginal likelihood $\pT(\bx)$ in DLVMs. Traditional inference methods are relatively expensive. Such methods, for example, often require a per-datapoint optimization loop, or yield bad posterior approximations. We would like to avoid such expensive procedures. 

Likewise, the posterior over the parameters of (directed models parameterized with) neural networks, $p(\bT|\mathcal{D})$, is generally intractable to compute exactly, and requires approximate inference techniques.

\chapter{Variational Autoencoders}
\label{chap:vaes}

In this chapter we explain the basics of variational autoencoders (VAEs).

\section{Encoder or Approximate Posterior}
\label{sec:encoder}

In the previous chapter, we introduced deep latent-variable models (DLVMs), and the problem of estimating the log-likelihood and posterior distributions in such models. The framework of variational autoencoders (VAEs) provides a computationally efficient way for optimizing DLVMs jointly with a corresponding inference model using SGD.

To turn the DLVM's intractable posterior inference and learning problems into tractable problems, we introduce a parametric \emph{inference model} $\qP(\bz|\bx)$. This model is also called an \emph{encoder} or \emph{recognition model}. With $\bphi$ we indicate the parameters of this inference model, also called the \emph{variational parameters}. We optimize the variational parameters $\bphi$ such that:
\begin{align}
\qP(\bz|\bx) \approx \pT(\bz|\bx)
\end{align}
As we will explain, this approximation to the posterior help us optimize the marginal likelihood.

Like a DLVM, the inference model can be (almost) any directed graphical model:
\begin{align}
\qP(\bz|\bx) = \qP(\bz_1,...,\bz_M | \bx) = \prod_{j=1}^M \qP(\bz_j | Pa(\bz_j), \bx) 
\end{align}
where $Pa(\bz_j)$ is the set of parent variables of variable $\bz_j$ in the directed graph. And also similar to a DLVM, the distribution $\qP(\bz|\bx)$ can be parameterized using deep neural networks. In this case, the variational parameters $\bphi$ include the weights and biases of the neural network. For example:
\begin{align}
(\bmu, \log \bsigma) &= \EncoderNeuralNet(\bx)\\
\qP(\bz|\bx) &= \mathcal{N}(\bz ; \bmu, \text{diag}(\bsigma))
\end{align}
Typically, we use a single encoder neural network to perform posterior inference over all of the datapoints in our dataset. This can be contrasted to more traditional variational inference methods where the variational parameters are not shared, but instead separately and iteratively optimized per datapoint. The strategy used in VAEs of sharing variational parameters across datapoints is also called \emph{amortized variational inference}~\citep{gershman2014amortized}. With amortized inference we can avoid a per-datapoint optimization loop, and leverage the efficiency of SGD.

\section{Evidence Lower Bound (ELBO)}
\label{sec:elbo}

\begin{figure}
	\centering
	\includegraphics[width=0.90\linewidth]{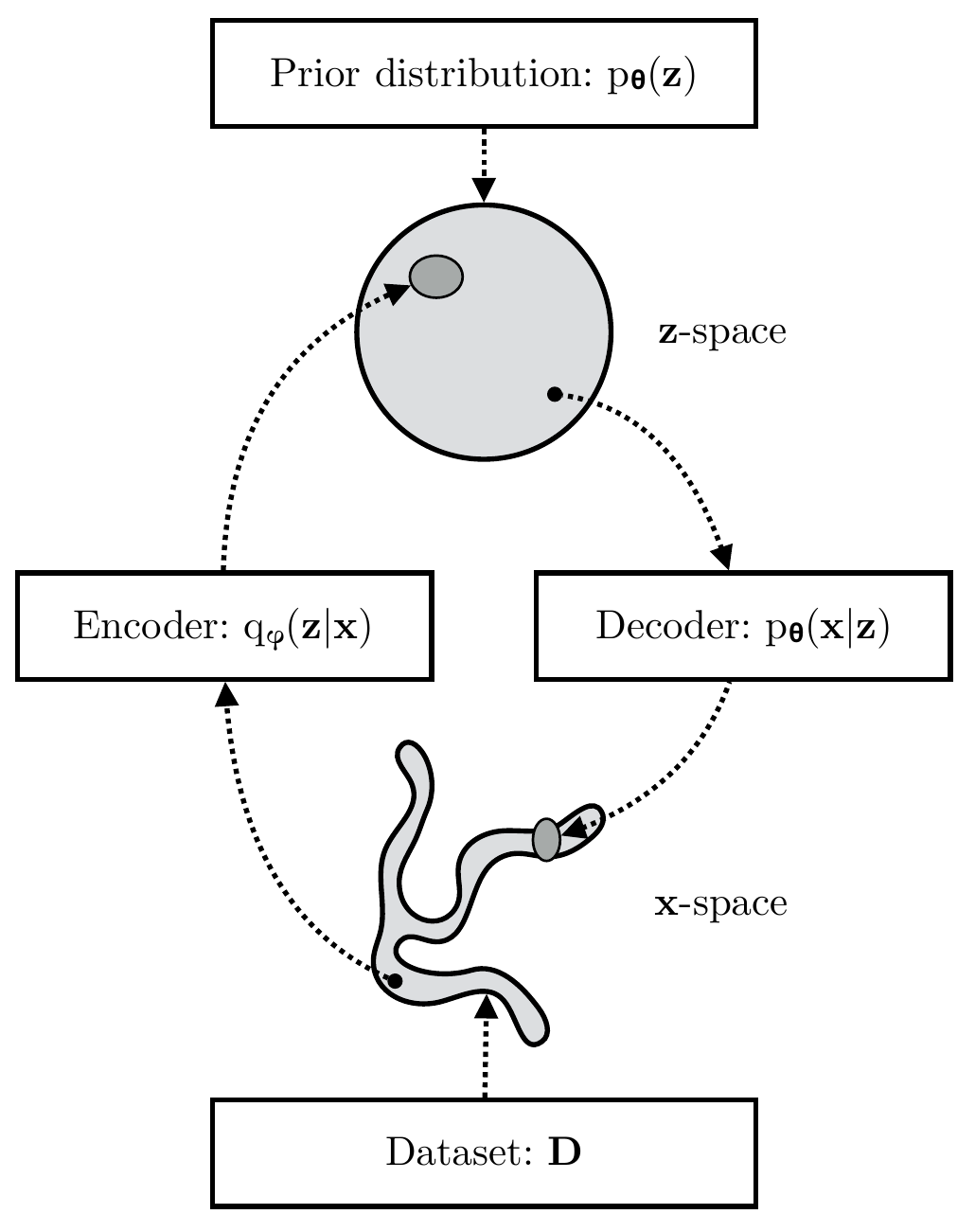}
	\caption{A VAE learns stochastic mappings between an observed $\bx$-space, whose empirical distribution $\qD(\bx)$ is typically complicated, and a latent $\bz$-space, whose distribution can be relatively simple (such as spherical, as in this figure). The generative model learns a joint distribution $\pT(\bx,\bz)$ that is often (but not always) factorized as $\pT(\bx,\bz) = \pT(\bz) \pT(\bx|\bz)$, with a prior distribution over latent space $\pT(\bz)$, and a stochastic decoder $\pT(\bx|\bz)$. The stochastic encoder $\qP(\bz|\bx)$, also called \emph{inference model}, approximates the true but intractable posterior $\pT(\bz|\bx)$ of the generative model.}
	\label{fig:xspacezspace}
\end{figure}

The optimization objective of the variational autoencoder, like in other variational methods, is the \emph{evidence lower bound}, abbreviated as ELBO. An alternative term for this objective is \emph{variational lower bound}. Typically, the ELBO is derived through Jensen's inequality. Here we will use an alternative derivation that avoids Jensen's inequality, providing greater insight about its tightness.

For any choice of inference model $\qP(\bz|\bx)$, including the choice of variational parameters $\bphi$, we have:
\begin{align}
\log \pT(\bx)
&= \Exp{\qP(\bz|\bx)}{ \log \pT(\bx) }\\
&= \Exp{\qP(\bz|\bx)}{ \log \left[ \frac{\pT(\bx,\bz)}{\pT(\bz|\bx)} \right] } \\
&= \Exp{\qP(\bz|\bx)}{ \log \left[ \frac{\pT(\bx,\bz)}{\qP(\bz|\bx)} \frac{\qP(\bz|\bx)}{\pT(\bz|\bx)} \right] } \\
&= \underbrace{
	\Exp{\qP(\bz|\bx)}{ \log \left[ \frac{\pT(\bx,\bz)}{\qP(\bz|\bx)} \right] }
}_{\substack{= \ELBO(\bx) \\ \text{\;\;(ELBO)}}}
+ \underbrace{\Exp{\qP(\bz|\bx)}{ \log \left[ \frac{\qP(\bz|\bx)}{\pT(\bz|\bx)} \right] }
}_{= D_{KL}(\qP(\bz|\bx)||\pT(\bz|\bx))}
\label{eq:elbokl}
\end{align}
The second term in eq.~\eqref{eq:elbokl} is the Kullback-Leibler (KL) divergence between $\qP(\bz|\bx)$ and $\pT(\bz|\bx)$, which is non-negative:
\begin{align}
D_{KL}(\qP(\bz|\bx)||\pT(\bz|\bx)) \geq 0
\end{align}
and zero if, and only if, $\qP(\bz|\bx)$ equals the true posterior distribution.

The first term in eq.~\eqref{eq:elbokl} is the \emph{variational lower bound}, also called the \emph{evidence lower bound} (ELBO):
\begin{align}
\ELBO(\bx) = \Exp{\qP(\bz|\bx)}{ \log \pT(\bx,\bz) - \log \qP(\bz|\bx) }
\label{eq:elbo}
\end{align}
Due to the non-negativity of the KL divergence, the ELBO is a lower bound on the log-likelihood of the data.
\begin{align}
\ELBO(\bx)
&= \log \pT(\bx) - D_{KL}(\qP(\bz|\bx)||\pT(\bz|\bx)) \label{eq:elbo2}\\
&\leq \log \pT(\bx)
\end{align}
So, interestingly, the KL divergence $D_{KL}(\qP(\bz|\bx)||\pT(\bz|\bx))$ determines two 'distances': 
\begin{enumerate}
	\item By definition, the KL divergence of the approximate posterior from the true posterior;
	\item The gap between the ELBO $\ELBO(\bx)$ and the marginal likelihood $\log \pT(\bx)$; this is also called the \emph{tightness} of the bound. The better $\qP(\bz|\bx)$ approximates the true (posterior) distribution $\pT(\bz|\bx)$, in terms of the KL divergence, the smaller the gap.
\end{enumerate}

\subsection{Two for One}
By looking at equation~\ref{eq:elbo2}, it can be understood that maximization of the ELBO $\ELBO(\bx)$ w.r.t. the parameters $\bT$ and $\bphi$, will concurrently optimize the two things we care about:
\begin{enumerate}
	\item It will approximately maximize the marginal likelihood $\pT(\bx)$. This means that our generative model will become better.
	\item It will minimize the KL divergence of the approximation $\qP(\bz|\bx)$ from the true posterior $\pT(\bz|\bx)$, so $\qP(\bz|\bx)$ becomes better. 
\end{enumerate}

\begin{figure}[t]
	\centering
	\includegraphics[width=0.90\linewidth]{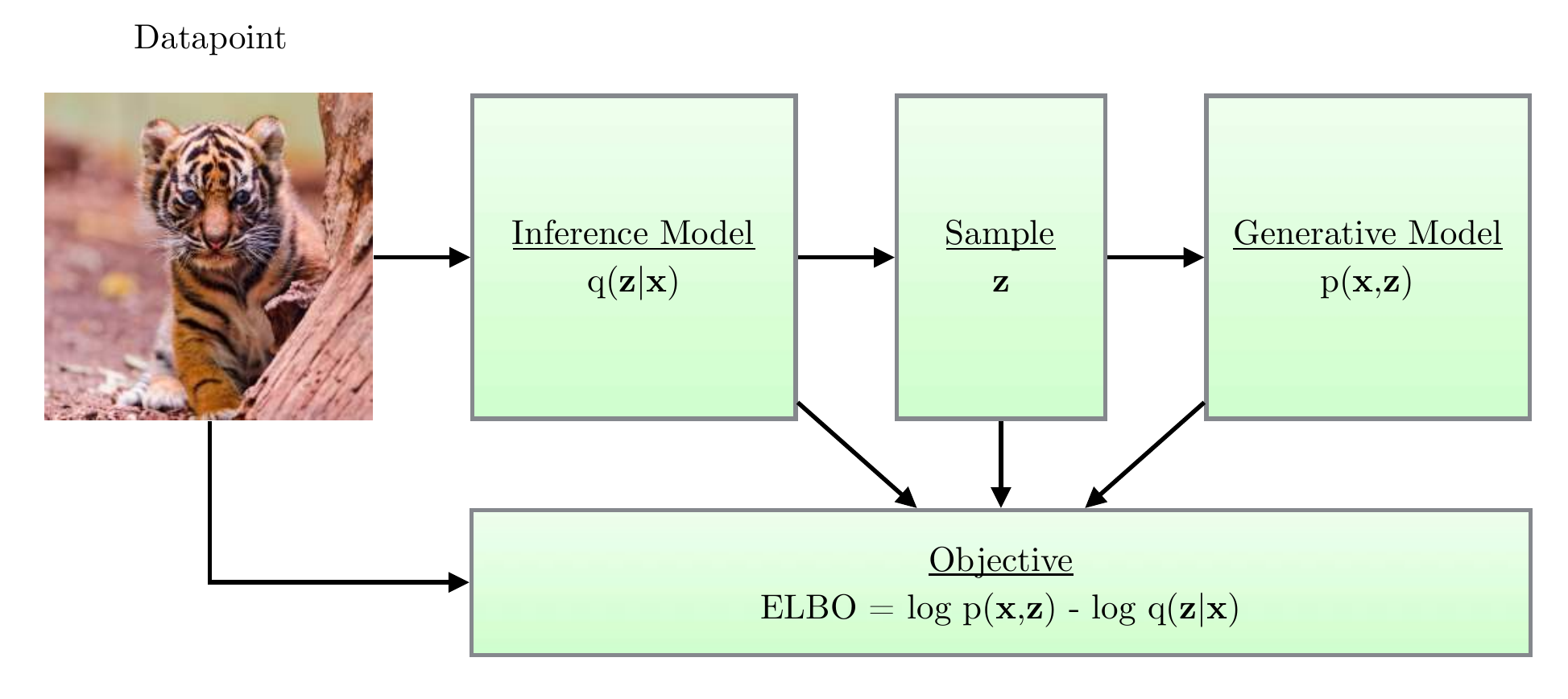}
	\caption{Simple schematic of computational flow in a variational autoencoder.}
	\label{fig:vaecartoon}
\end{figure}

\section{Stochastic Gradient-Based Optimization of the ELBO}

An important property of the ELBO, is that it allows \emph{joint} optimization w.r.t. all parameters ($\bphi$ and $\bT$) using stochastic gradient descent (SGD). We can start out with random initial values of $\bphi$ and $\bT$, and stochastically optimize their values until convergence.

Given a dataset with i.i.d. data, the ELBO objective is the sum (or average) of individual-datapoint ELBO's:
\begin{align}
\ELBO(\mathcal{D}) = \sum_{\bx \in \mathcal{D}} \ELBO(\bx)
\end{align}
The individual-datapoint ELBO, and its gradient $\nabla_{\bT,\bphi} \ELBO(\bx)$ is, in general, intractable. However, good unbiased estimators $\tilde{\nabla}_{\bT,\bphi} \ELBO(\bx)$ exist, as we will show, such that we can still perform minibatch SGD.

Unbiased gradients of the ELBO w.r.t. the generative model parameters $\bT$ are simple to obtain:
\begin{align}
\nabla_{\bT} \ELBO(\bx) &= \nabla_{\bT} \Exp{\qP(\bz|\bx)}{ \log \pT(\bx,\bz) - \log \qP(\bz|\bx) }\\
&= \Exp{\qP(\bz|\bx)}{\nabla_{\bT}( \log \pT(\bx,\bz) - \log \qP(\bz|\bx)) }
\label{eq:nablabtelbo2}\\
&\simeq \nabla_{\bT}( \log \pT(\bx,\bz) - \log \qP(\bz|\bx))\label{eq:nablabtelbo3}\\
&= \nabla_{\bT}( \log \pT(\bx,\bz))\label{eq:nablabtelbo}
\end{align}
The last line (eq. \eqref{eq:nablabtelbo}) is a simple Monte Carlo estimator of the second line (eq. \eqref{eq:nablabtelbo2}), where $\bz$ in the last two lines (eq. \eqref{eq:nablabtelbo3} and eq. \eqref{eq:nablabtelbo}) is a random sample from $\qP(\bz|\bx)$.

Unbiased gradients w.r.t. the \emph{variational} parameters $\bphi$ are more difficult to obtain, since the ELBO's expectation is taken w.r.t. the distribution $\qP(\bz|\bx)$, which is a function of $\bphi$. I.e., in general:
\begin{align}
\nabla_{\bphi} \ELBO(\bx) &= \nabla_{\bphi} \Exp{\qP(\bz|\bx)}{ \log \pT(\bx,\bz) - \log \qP(\bz|\bx) }\\
&\neq \Exp{\qP(\bz|\bx)}{\nabla_{\bphi}( \log \pT(\bx,\bz) - \log \qP(\bz|\bx)) }\end{align}

In the case of continuous latent variables, we can use a reparameterization trick for computing unbiased estimates of $\nabla_{\bT,\bphi} \ELBO(\bx)$, as we will now discuss. This stochastic estimate allows us to optimize the ELBO using SGD; see algorithm~\ref{algorithm:elbooptim}. See section~\ref{sec:reinforce} for a discussion of variational methods for discrete latent variables.

\begin{algorithm}[t]
	\SetKwInOut{Input}{input}
	\SetKwInOut{Output}{output}
	\KwData{\\
		\hspace{5mm}$\mathcal{D}$: Dataset\\ 
		\hspace{5mm}$\qP(\bz|\bx)$: Inference model\\
		\hspace{5mm}$\pT(\bx, \bz)$: Generative model\\		
	}
	\KwResult{\\
		\hspace{5mm}$\bT, \bphi$: Learned parameters\\
	}
	\BlankLine
	$(\bT, \bphi) \leftarrow \text{Initialize parameters}$\\
	\While{SGD not converged}{
		$\mathcal{M} \sim \mathcal{D}$ (Random minibatch of data)\\
		$\beps \sim p(\beps)$ (Random noise for every datapoint in $\mathcal{M}$)\\
		Compute $\TELBO(\mathcal{M},\beps)$ and its gradients $\nabla_{\bT,\bphi}\TELBO(\mathcal{M},\beps)$\\
		Update $\bT$ and $\bphi$ using SGD optimizer\\
	}
	\caption{Stochastic optimization of the ELBO. Since noise originates from both the minibatch sampling and sampling of $p(\beps)$, this is a doubly stochastic optimization procedure. We also refer to this procedure as the \emph{Auto-Encoding Variational Bayes} (AEVB) algorithm.}
	\label{algorithm:elbooptim}
\end{algorithm}

\section{Reparameterization Trick}
\label{sec:reparameterization_trick}

For continuous latent variables and a differentiable encoder and generative model, the ELBO can be straightforwardly differentiated w.r.t. both $\bphi$ and $\bT$ through a change of variables, also called the \emph{reparameterization trick} (\cite{kingma2013auto} and \cite{rezende2014stochastic}).

\subsection{Change of variables}
First, we express the random variable $\bz \sim \qP(\bz|\bx)$ as some differentiable (and invertible) transformation of another random variable $\beps$, given $\bz$ and $\bphi$:
\begin{align}
\bz = \bg(\beps,\bphi,\bx)
\end{align}
where the distribution of random variable $\beps$ is independent of $\bx$ or $\bphi$.

\subsection{Gradient of expectation under change of variable}
\begin{figure}
	\centering
	\includegraphics[width=0.9\linewidth]{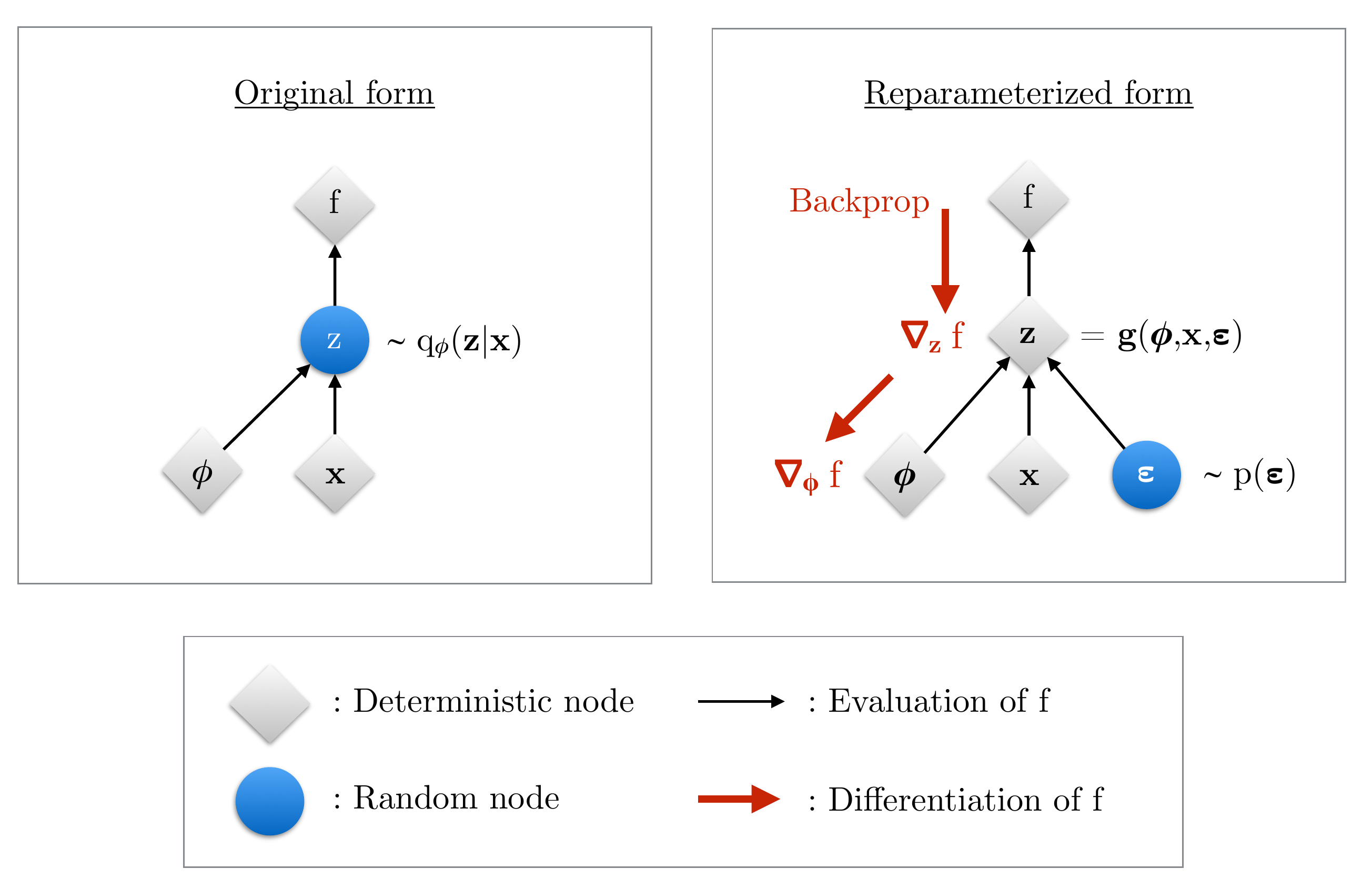}
	\caption{Illustration of the reparameterization trick. The variational parameters $\bphi$ affect the objective $f$ through the random variable $\bz \sim \qP(\bz|\bx)$. We wish to compute gradients $\nabla_{\bphi} f$ to optimize the objective with SGD. In the original form (left), we cannot differentiate $f$ w.r.t. $\bphi$, because we cannot directly backpropagate gradients through the random variable $\bz$. We can 'externalize' the randomness in $\bz$ by re-parameterizing the variable as a deterministic and differentiable function of $\bphi$, $\bx$, and a newly introduced random variable $\beps$. This allows us to 'backprop through $\bz$', and compute gradients $\nabla_{\bphi} f$.}
	\label{fig:reparameterization}
\end{figure}

\begin{figure}[t]
	\centering
	\begin{subfigure}[]{0.33\textwidth}
		\centering
		\includegraphics[width=1\textwidth]{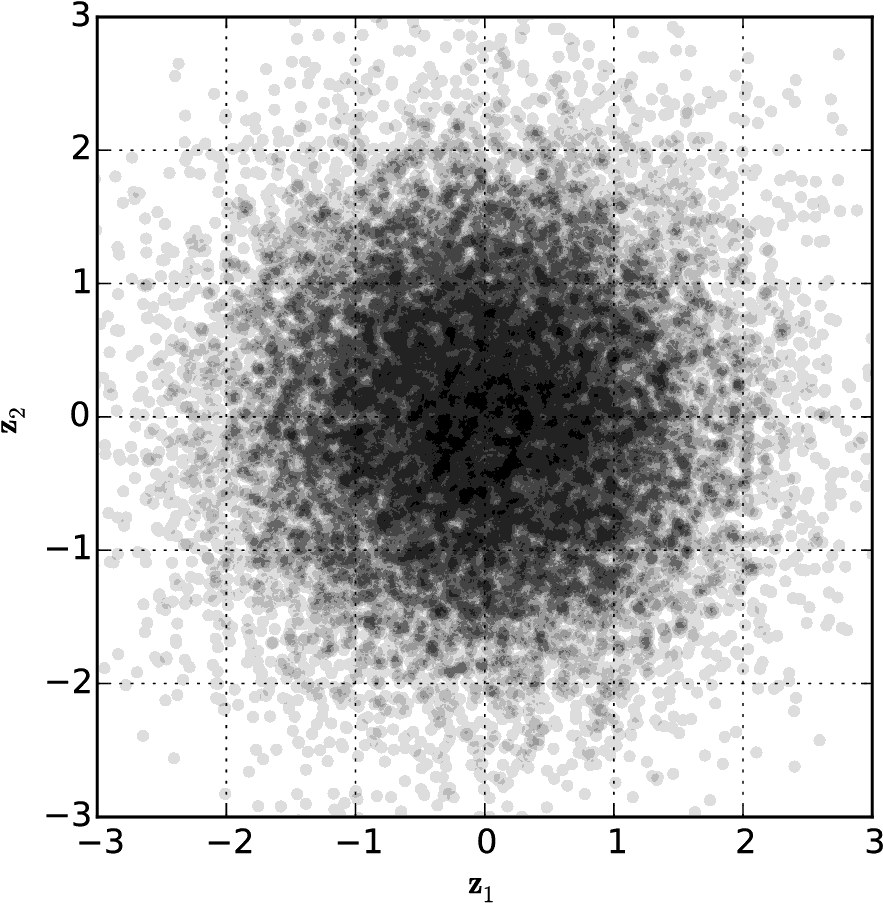}
		\caption{Prior distribution}
	\end{subfigure}%
	\begin{subfigure}[]{0.33\textwidth}
		\centering
		\includegraphics[width=1\textwidth]{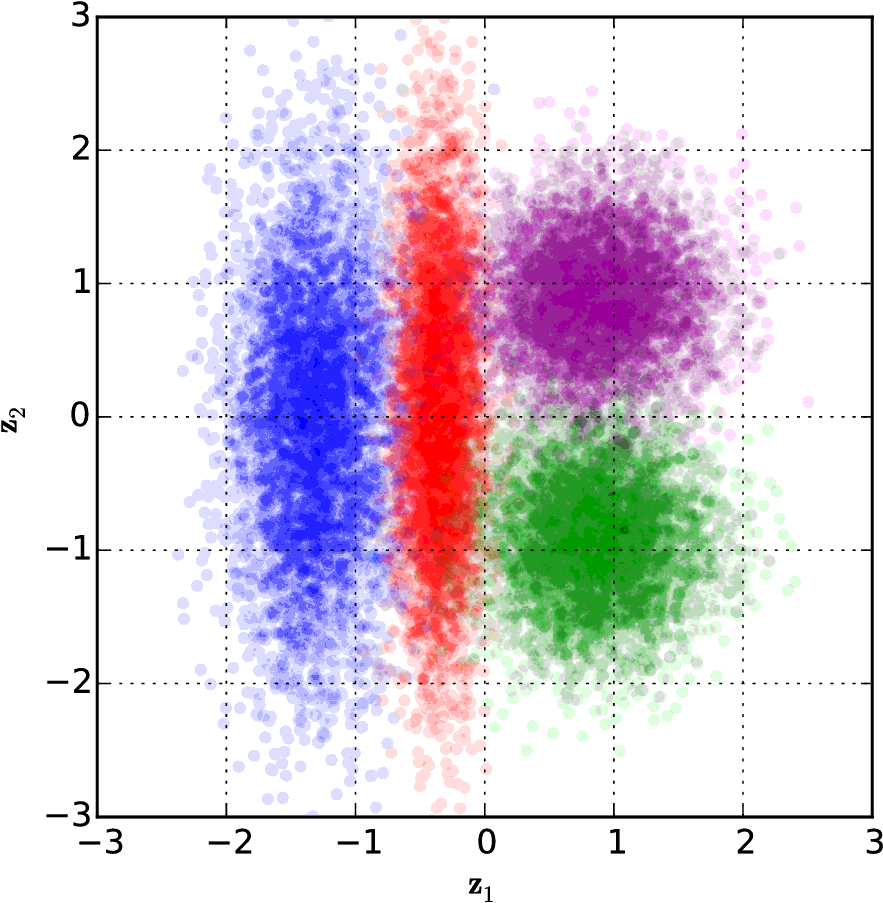}
		\caption{Factorized posteriors}
	\end{subfigure}%
	\begin{subfigure}[]{0.33\textwidth}
		\centering
		\includegraphics[width=1\textwidth]{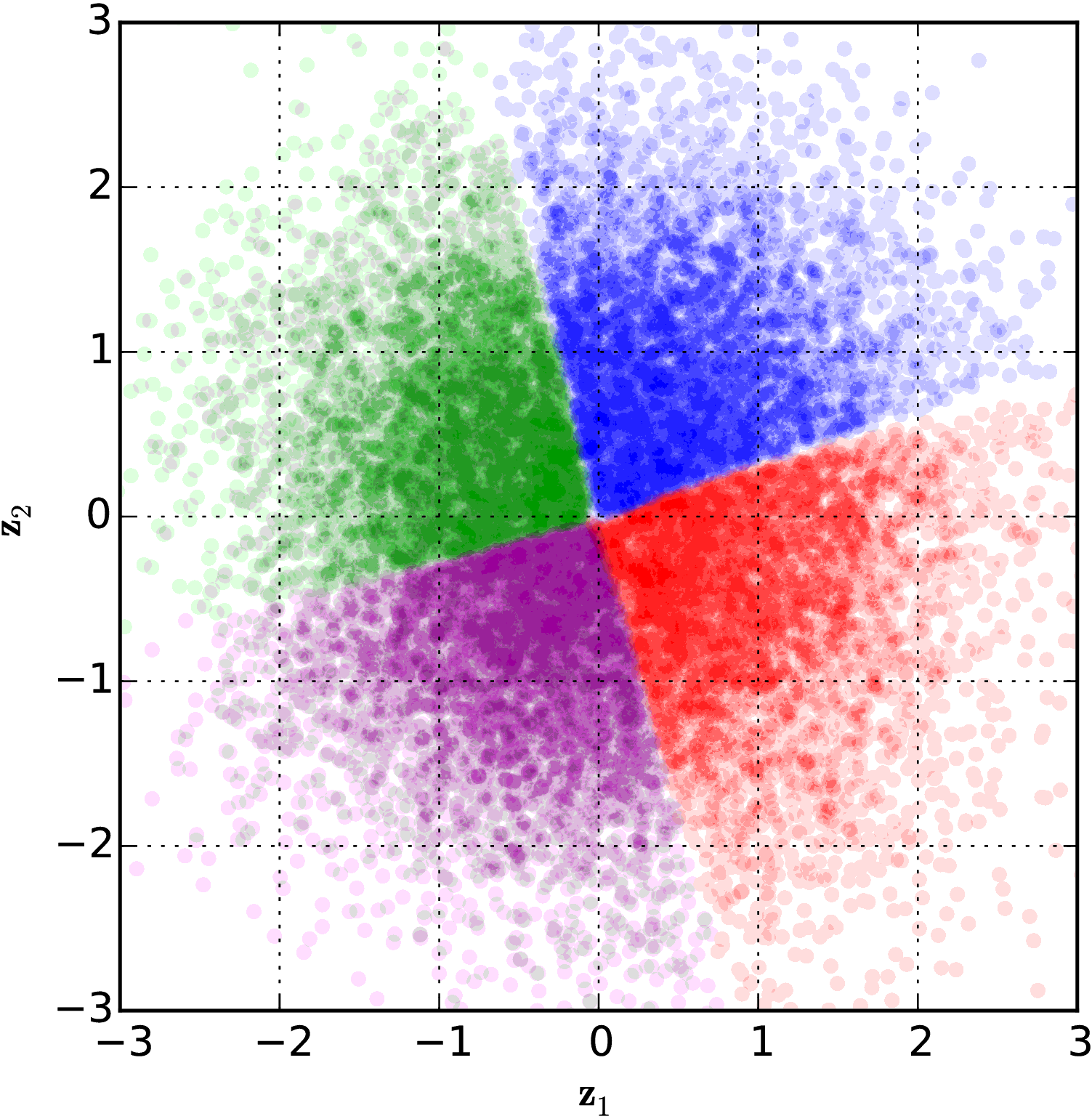}
		\caption{IAF posteriors}
	\end{subfigure}%
	\caption{Best viewed in color. We fitted a variational autoencoder (VAE) with a spherical Gaussian prior, and with factorized Gaussian posteriors \textbf{(b)} or inverse autoregressive flow (IAF) posteriors \textbf{(c)} to a toy dataset with four datapoints. Each colored cluster corresponds to the posterior distribution of one datapoint. IAF greatly improves the flexibility of the posterior distributions, and allows for a much better fit between the posteriors and the prior.}
	\label{fig:xor}
\end{figure}

Given such a change of variable, expectations can be rewritten in terms of $
\beps$, 
\begin{align}
\Exp{\qP(\bz|\bx)}{f(\bz)} = \Exp{p(\beps)}{f(\bz)}
\end{align}
where $\bz = \bg(\beps,\bphi,\bx)$.
and the expectation and gradient operators become commutative, and we can form a simple Monte Carlo estimator:
\begin{align}
\nabla_{\bphi} \Exp{\qP(\bz|\bx)}{f(\bz)} 
&= \nabla_{\bphi} \Exp{p(\beps)}{f(\bz)} \\
&=  \Exp{p(\beps)}{\nabla_{\bphi} f(\bz)}\\
&\simeq \nabla_{\bphi} f(\bz)
\end{align}
where in the last line, $\bz = \bg(\bphi,\bx,\beps)$ with random noise sample $\beps \sim p(\beps)$.
See figure~\ref{fig:reparameterization} for an illustration and further clarification, and figure~\ref{fig:xor} for an illustration of the resulting posteriors for a 2D toy problem.

\subsection{Gradient of ELBO}

Under the reparameterization, we can replace an expectation w.r.t. $\qP(\bz|\bx)$ with one w.r.t. $p(\beps)$. The ELBO can be rewritten as:
\begin{align}
\ELBO(\bx) &= \Exp{\qP(\bz|\bx)}{ \log \pT(\bx,\bz) - \log \qP(\bz|\bx) }\\
&= \Exp{p(\beps)}{\log \pT(\bx,\bz) - \log \qP(\bz|\bx) }
\end{align}
where $\bz = g(\beps,\bphi,\bx)$.

As a result we can form a simple Monte Carlo estimator $\TELBO(\bx)$ of the individual-datapoint ELBO where we use a single noise sample $\beps$ from $p(\beps)$:
\begin{align}
\beps &\sim  p(\beps)\\
\bz &= \bg(\bphi,\bx,\beps)\\
\TELBO(\bx) &=  \log \pT(\bx,\bz) - \log \qP(\bz|\bx)
\end{align}

This series of operations can be expressed as a symbolic graph in software like TensorFlow, and effortlessly differentiated w.r.t. the parameters $\bT$ and $\bphi$. The resulting gradient $\nabla_{\bphi} \TELBO(\bx)$ is used to optimize the ELBO using minibatch SGD. See algorithm~\ref{algorithm:elbooptim}. This algorithm was originally referred to as the Auto-Encoding Variational Bayes (AEVB) algorithm by ~\cite{kingma2013auto}. More generally, the reparameterized ELBO estimator was referred to as the Stochastic Gradient Variational Bayes (SGVB) estimator. This estimator can also be used to estimate a posterior over the model parameters, as explained in the appendix of ~\citep{kingma2013auto}.

\subsubsection{Unbiasedness}
This gradient is an unbiased estimator of the exact single-datapoint ELBO gradient; when averaged over noise $\beps \sim p(\beps)$, this gradient equals the single-datapoint ELBO gradient:
\begin{align}
\Exp{p(\beps)}{\nabla_{\bT,\bphi} \TELBO(\bx;\beps)}
&= \Exp{p(\beps)}{\nabla_{\bT,\bphi}  (\log \pT(\bx,\bz) - \log \qP(\bz|\bx)) }\\
&= \nabla_{\bT,\bphi} ( \Exp{p(\beps)}{\log \pT(\bx,\bz) - \log \qP(\bz|\bx) } )\\
&= \nabla_{\bT,\bphi} \ELBO(\bx) 
\end{align}

\subsection{Computation of $\log \qP(\bz|\bx)$}
\label{sec:posteriordensity}

Computation of the (estimator of) the ELBO requires computation of the density $\log \qP(\bz|\bx)$, given a value of $\bx$, and given a value of $\bz$ or equivalently $\beps$. This log-density is a simple computation, as long as we choose the right transformation $\bg()$.

Note that we typically know the density $p(\beps)$, since this is the density of the chosen noise distribution. As long as $\bg(.)$ is an invertible function, the densities of $\beps$ and $\bz$ are related as:
\begin{align}
\log \qP(\bz|\bx) = \log p(\beps) - \log \dP(\bx,\beps)
\end{align}
where the second term is the log of the absolute value of the determinant of the Jacobian matrix $(\partial \bz / \partial \beps)$:
\begin{align}
\log \dP(\bx,\beps) =  \log \left| \det \left(\frac{\partial \bz}{ \partial \beps} \right) \right|
\end{align}
We call this the log-determinant of the transformation from $\beps$ to $\bz$.  We use the notation $\log \dP(\bx,\beps)$ to make explicit that this log-determinant, similar to $\bg()$, is a function of $\bx$, $\beps$ and $\bphi$. The Jacobian matrix contains all first derivatives of the transformation from $\beps$ to $\bz$:
\begin{align}
\frac{\partial \bz}{\partial \beps}
= \frac{\partial (z_1,...,z_k)}{\partial (\epsilon_1,...,\epsilon_k)}
= \begin{pmatrix}
\frac{\partial z_1}{\partial \epsilon_1} & \cdots & \frac{\partial z_1}{\partial \epsilon_k} \\
\vdots  & \ddots & \vdots  \\
\frac{\partial z_k}{\partial \epsilon_1} & \cdots & \frac{\partial z_k}{\partial \epsilon_k} \\
\end{pmatrix}
\end{align}
As we will show, we can build very flexible transformations $\bg()$ for which $\log \dP(\bx,\beps)$ is simple to compute, resulting in highly flexible inference models $\qP(\bz|\bx)$.

\section{Factorized Gaussian posteriors}

A common choice is a simple factorized Gaussian encoder\\$\qP(\bz|\bx) = \mathcal{N}(\bz; \bmu, \text{diag}(\bsigma^2))$:
\begin{align}
(\bmu, \log \bsigma) &= \EncoderNeuralNet(\bx)\\
\qP(\bz|\bx) &= \prod_i \qP(z_i|\bx) = \prod_i \mathcal{N}(z_i; \mu_i, \sigma_i^2)
\end{align}
where $\mathcal{N}(z_i; \mu_i, \sigma_i^2)$ is the PDF of the univariate Gaussian distribution. After reparameterization, we can write:
\begin{align}
\beps &\sim \mathcal{N}(0,\bI)\\
(\bmu, \log \bsigma) &= \EncoderNeuralNet(\bx)\\
\bz &= \bmu + \bsigma \odot \beps
\end{align}
where $\odot$ is the element-wise product. The Jacobian of the transformation from $\beps$ to $\bz$ is:
\begin{align}
\frac{\partial \bz}{\partial \beps} = \text{diag}(\bsigma),
\end{align}
i.e. a diagonal matrix with the elements of $\bsigma$ on the diagonal. The determinant of a diagonal (or more generally, triangular) matrix is the product of its diagonal terms. The log determinant of the Jacobian is therefore:
\begin{align}
\log \dP(\bx,\beps) = \log \left| \det \left(\frac{\partial \bz}{\partial \beps}\right) \right| = \sum_i \log \sigma_i
\label{eq:factgausslogdet}
\end{align}
and the posterior density is:
\begin{align}
\log \qP(\bz|\bx)
&= \log p(\beps) - \log \dP(\bx,\beps)\\
&= \sum_i \log \mathcal{N}(\epsilon_i; 0, 1) - \log \sigma_i
\end{align}
when $z=g(\beps,\phi,\bx)$.

\subsection{Full-covariance Gaussian posterior}
The factorized Gaussian posterior can be extended to a Gaussian with full covariance:
\begin{align}
\qP(\bz|\bx) = \mathcal{N}(\bz; \bmu, \boldsymbol{\Sigma})
\end{align}
A reparameterization of this distribution is given by:
\begin{align}
\beps &\sim \mathcal{N}(0,\bI)\\
\bz &= \bmu + \bL \beps
\end{align}
where $\bL$ is a lower (or upper) triangular matrix, with non-zero entries on the diagonal. The off-diagonal elements define the correlations (covariances) of the elements in $\bz$.

The reason for this parameterization of the full-covariance Gaussian, is that the Jacobian determinant is remarkably simple. The Jacobian in this case is trivial: $\frac{\partial \bz}{\partial \beps} = \bL$. Note that the determinant of a triangular matrix is the product of its diagonal elements. Therefore, in this parameterization:
\begin{align}
\log | \det(\frac{\partial \bz}{\partial \beps}) | &= \sum_i \log |L_{ii}|
\end{align}
And the log-density of the posterior is:
\begin{align}
\log \qP(\bz|\bx) &= \log p(\beps) - \sum_i \log |L_{ii}|
\end{align}

This parameterization corresponds to the Cholesky decomposition $\boldsymbol{\Sigma} = \bL \bL^T$ of the covariance of $\bz$:
\begin{align}
\boldsymbol{\Sigma}
&= \Exp{}{(\bz - \Exp{}{\bz|}) (\bz - \Exp{}{\bz|})^T}\\
&= \Exp{}{\bL \beps (\bL \beps)^T} = \bL \Exp{}{\beps \beps^T} \bL^T\\ 
&= \bL \bL^T
\end{align}
Note that $\Exp{}{\beps \beps^T} = \bI$ since $\beps \sim \mathcal{N}(0, \bI)$.

One way to build a matrix $\bL$ with the desired properties, namely triangularity and non-zero diagonal entries, is by constructing it as follows:
\begin{align}
(\bmu, \log \bsigma, \bL') \leftarrow \EncoderNeuralNet(\bx)\\
\bL \leftarrow \bL_{mask} \odot \bL' + \text{diag}(\bsigma)
\end{align}
and then proceeding with $\bz = \bmu + \bL \beps$ as described above. $\bL_{mask}$ is a masking matrix with zeros on and above the diagonal, and ones below the diagonal. Note that due to the masking $\bL$, the Jacobian matrix $(\partial \bz / \partial \beps)$ is triangular with the values of $\bsigma$ on the diagonal. The log-determinant is therefore identical to the factorized Gaussian case:
\begin{align}
\log \left| \det \left(\frac{\partial \bz}{\partial \beps}\right) \right| = \sum_i \log \sigma_i
\label{eq:fullgausslogdet}
\end{align}
More generally, we can replace $\bz = \bL \beps + \bmu$ with a chain of (differentiable and nonlinear) transformations; as long as the Jacobian of each step in the chain is triangular with non-zero diagonal entries, the log determinant remains simple. This principle is used by \emph{inverse autoregressive flow} (IAF) as explored by ~\cite{kingma2016improving} and discussed in chapter ~\ref{chap:advanced_q}.

\begin{algorithm}
	\SetKwFunction{EncoderNN}{EncoderNN}
	\SetKwFunction{AutoregressiveNN}{AutoregressiveNN}
	\SetKwFunction{sigmoid}{sigmoid}
	\SetKwFunction{dosum}{sum}
	\SetKwFunction{sigmoid}{sigmoid}
	\SetKwInOut{Input}{input}
	\SetKwInOut{Output}{output}
	\KwData{\\
		\hspace{5mm}$\bx$: a datapoint, and optionally other conditioning information\\
		\hspace{5mm}$\beps$: a random sample from $p(\beps) = \mathcal{N}(0,\bI)$\\
		\hspace{5mm}$\bT$: Generative model parameters\\
		\hspace{5mm}$\bphi$: Inference model parameters\\
		\hspace{5mm}$\qP(\bz|\bx)$: Inference model\\
		\hspace{5mm}$\pT(\bx, \bz)$: Generative model\\		
	}
	\KwResult{\\
		\hspace{5mm}$\tilde{\mathcal{L}}$: unbiased estimate of the single-datapoint ELBO $\ELBO(\bx)$\\
	}
	\BlankLine
	$(\bmu, \log \bsigma, \bL') \leftarrow \EncoderNeuralNet(\bx)$\\
	$\bL \leftarrow \bL_{mask} \odot \bL' + \text{diag}(\bsigma)$\\
	$\bepsilon \sim \mathcal{N}(0, \bI)$\\
	$\bz \leftarrow \bL \bepsilon + \bmu$\\
	$\tilde{\mathcal{L}}_\text{logqz} \leftarrow - \sum_i (\tfrac{1}{2} (\epsilon_i^2 + \log(2 \pi) + \log \sigma_i))_i$ \Comment{$= \qP(\bz|\bx)$}\\
	$\tilde{\mathcal{L}}_\text{logpz} \leftarrow -\sum_i(\tfrac{1}{2}(z_i^2 + \log(2 \pi)))$\Comment{$= \pT(\bz)$}\\
	$\bp \leftarrow \DecoderNeuralNet(\bz)$\\
	$\tilde{\mathcal{L}}_\text{logpx} \leftarrow \sum_i(x_i \log p_i + (1-x_i) \log (1-p_i))$ \Comment{$= \pT(\bx|\bz)$}\\
	$\tilde{\mathcal{L}} = \tilde{\mathcal{L}}_\text{logpx} + \tilde{\mathcal{L}}_\text{logpz} - \tilde{\mathcal{L}}_\text{logqz}$
	\caption{Computation of unbiased estimate of single-datapoint ELBO for example VAE with a full-covariance Gaussian inference model and a factorized Bernoulli generative model. $\bL_{mask}$ is a masking matrix with zeros on and above the diagonal, and ones below the diagonal.
	}
	\label{algorithm:vae}
\end{algorithm}

\section{Estimation of the Marginal Likelihood}

After training a VAE, we can estimate the probability of data under the model using an \emph{importance sampling} technique, as originally proposed by ~\cite{rezende2014stochastic}. The marginal likelhood of a datapoint can be written as:
\begin{align}
\log \pT(\bx) = \log \Exp{\qP(\bz|\bx)}{\pT(\bx,\bz)/\qP(\bz|\bx)}
\end{align}
Taking random samples from $\qP(\bz|\bx)$, a Monte Carlo estimator of this is:
\begin{align}
\log \pT(\bx) \approx \log \frac{1}{L} \sum_{l=1}^L \pT(\bx,\bz^{(l)})/\qP(\bz^{(l)}|\bx)
\label{eq:importanceweightedestimator}
\end{align}
where each $\bz^{(l)} \sim \qP(\bz|\bx)$ is a random sample from the inference model. By making $L$ large, the approximation becomes a better estimate of the marginal likelihood, and in fact since this is a Monte Carlo estimator, for $L \to \infty$ this converges to the actual marginal likelihood.

Notice that when setting $L=1$, this equals the ELBO estimator of the VAE. We can also use the estimator of eq. ~\eqref{eq:importanceweightedestimator} as our objective function; this is the objective used in \emph{importance weighted autoencoders}~\citep{burda2015importance} (IWAE). In that paper, it was also shown that the objective has increasing tightness for increasing value of $L$. It was later shown by ~\cite{cremer2017reinterpreting} that the IWAE objective can be re-interpreted as an ELBO objective with a particular inference model. The downside of these approaches for optimizing a tighter bound, is that importance weighted estimates have notoriously bad scaling properties to high-dimensional latent spaces.

\section{Marginal Likelihood and ELBO as KL Divergences}
\label{sec:elbokl}

One way to improve the potential tightness of the ELBO, is increasing the flexibility of the generative model. This can be understood through a connection between the ELBO and the KL divergence.

\looseness=-1 With i.i.d. dataset $\mathcal{D}$ of size $N_\mathcal{D}$, the maximum likelihood criterion is: 
\begin{align}
\log \pT(\mathcal{D})
&= \frac{1}{N_\mathcal{D}} \sum_{\bx \in \mathcal{D}} \log \pT(\bx)\\
&= \Exp{\qD(\bx)}{ \log \pT(\bx) }
\end{align}
where $\qD(\bx)$ is the empirical (data) distribution, which is a mixture distribution:
\begin{align}
\qD(\bx) = \frac{1}{N} \sum_{i=1}^N \qD^{(i)}(\bx)
\label{eq:empiricaldist}
\end{align}
where each component $\qD^{(i)}(\bx)$ typically corresponds to a Dirac delta distribution centered at value $\bx^{(i)}$ in case of continuous data, or a discrete distribution with all probability mass concentrated at value $\bx^{(i)}$ in case of discrete data. The Kullback Leibler (KL) divergence between the data and model distributions, can be rewritten as the negative log-likelihood, plus a constant:
\begin{align}
D_{KL}(\qD(\bx)||\pT(\bx))
&= - \Exp{\qD(\bx)}{ \log \pT(\bx) } + \Exp{\qD(\bx)}{\log \qD(\bx) }\\
&= - \log \pT(\mathcal{D}) + \text{constant}
\label{eq:marglikkl}
\end{align}
where $\text{constant} = - \mathcal{H}(\qD(\bx))$. So minimization of the KL divergence above is equivalent to maximization of the data log-likelihood $\log \pT(\mathcal{D})$.

Taking the combination of the empirical data distribution $\qD(\bx)$ and the inference model, we get a joint distribution over data $\bx$ and latent variables $\bz$: $\qDP(\bx,\bz) = \qD(\bx)q(\bz|\bx)$.

\begin{figure}
	\centering
	\includegraphics[width=0.9\linewidth]{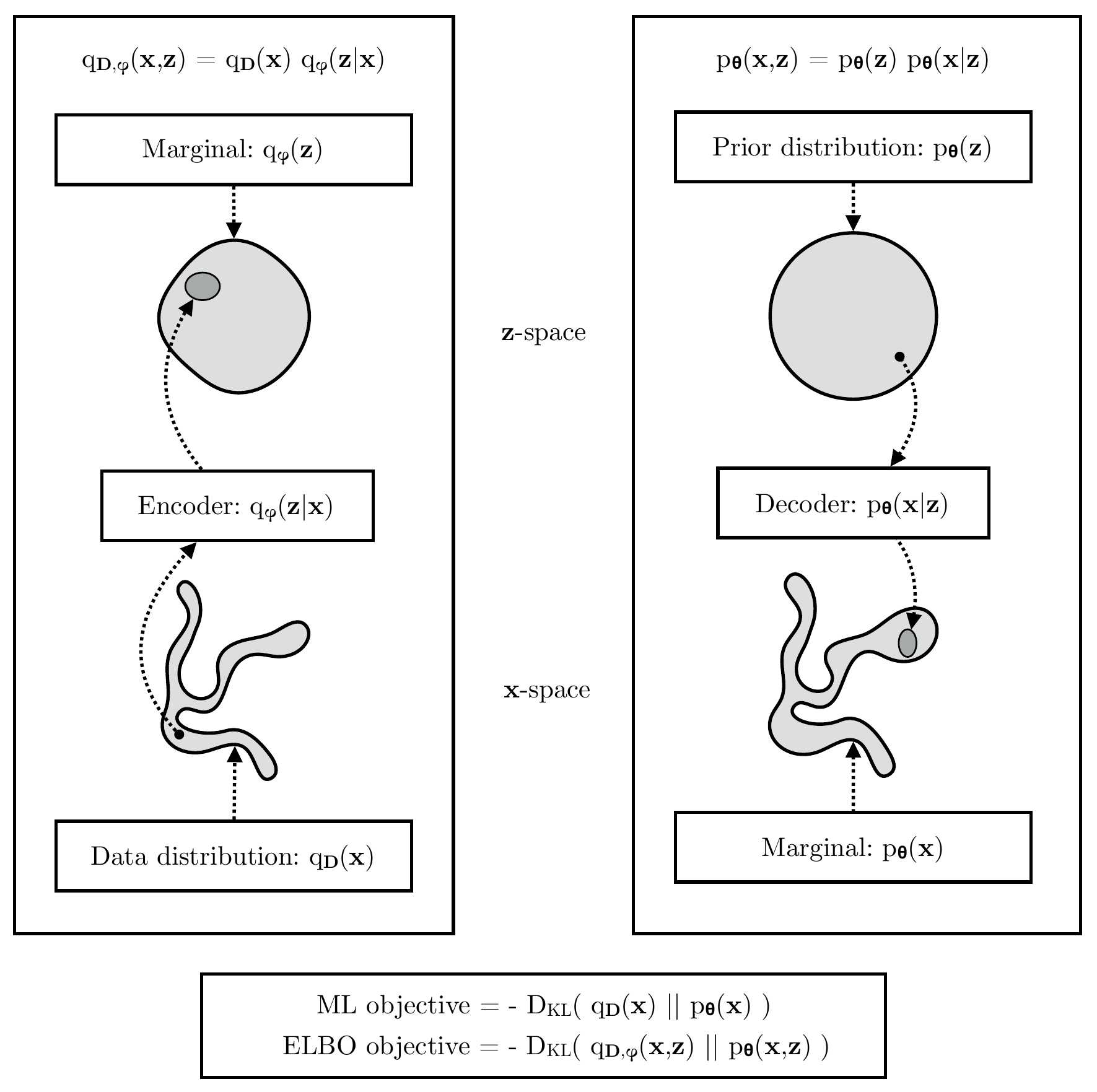}
	\caption{The maximum likelihood (ML) objective can be viewed as the minimization of $D_{KL}(\qDP(\bx) || \pT(\bx))$, while the ELBO objective can be viewed as the minimization of $D_{KL}(\qDP(\bx,\bz) || \pT(\bx,\bz))$, which upper bounds $D_{KL}(\qDP(\bx) || \pT(\bx))$. If a perfect fit is not possible, then $\pT(\bx,\bz)$ will typically end up with higher variance than $\qDP(\bx,\bz)$, because of the direction of the KL divergence.}
	\label{fig:klqxzpxz}
\end{figure}

The KL divergence of $\qDP(\bx,\bz)$ from $\pT(\bx,\bz)$ can be written as the negative ELBO, plus a constant:
\begin{align}
&D_{KL}(\qDP(\bx,\bz) || \pT(\bx,\bz))\\
&= - \Exp{\qD(\bx)}{\Exp{\qP(\bz|\bx)}{\log \pT(\bx,\bz) - \log \qP(\bz|\bx)} - \log \qD(\bx)}\\
&= - \ELBO(\mathcal{D}) + \text{constant}
\label{eq:elbokl2}
\end{align}
where $\text{constant} = - \mathcal{H}(\qD(\bx))$. So maximization of the ELBO, is equivalent to the minimization of this KL divergence $D_{KL}(\qDP(\bx,\bz) || \pT(\bx,\bz))$.
The relationship between the ML and ELBO objectives can be summarized in the following simple equation:
\begin{align}
&D_{KL}(\qDP(\bx,\bz) || \pT(\bx,\bz))\\
&= D_{KL}(\qD(\bx) || \pT(\bx)) + \Exp{\qD(\bx)}{D_{KL}(\qDP(\bz|\bx) || \pT(\bz|\bx))}\\
&\geq D_{KL}(\qD(\bx) || \pT(\bx))
\end{align}
One additional perspective is that the ELBO can be viewed as a maximum likelihood objective \emph{in an augmented space}. For some fixed choice of encoder $\qP(\bz|\bx)$, we can view the joint distribution $\pT(\bx,\bz)$ as an augmented empirical distribution over the original data $\bx$ and (stochastic) auxiliary features $\bz$ associated with each datapoint. The model $\pT(\bx,\bz)$ then defines a joint model over the original data, and the auxiliary features. See figure~\ref{fig:klqxzpxz}.

\section{Challenges}

\subsection{Optimization issues}

In our work, consistent with findings in~\citep{bowman2015generating} and \citep{sonderby2016train}, we found that stochastic optimization with the unmodified lower bound objective can gets stuck in an undesirable stable equilibrium. At the start of training, the likelihood term $\log p(\bx|\bz)$ is relatively weak, such that an initially attractive state is where $q(\bz|\bx) \approx p(\bz)$, resulting in a stable equilibrium from which it is difficult to escape. The solution proposed in~\citep{bowman2015generating} and \citep{sonderby2016train} is to use an optimization schedule where the weights of the latent cost $D_{KL}(q(\bz|\bx)||p(\bz))$ is slowly annealed from $0$ to $1$ over many epochs. 

An alternative proposed in~\citep{kingma2016improving} is the method of \emph{free bits}: a modification of the ELBO objective, that ensures that on average, a certain minimum number of bits of information are encoded per latent variable, or per group of latent variables.

The latent dimensions are divided into the $K$ groups. We then use the following minibatch objective, which ensures that using less than $\lambda$ nats of information per subset $j$ (on average per minibatch $\mathcal{M}$) is not advantageous:
\begin{align}
\widetilde{\mathcal{L}}_\lambda &= \Exp{\bx \sim \mathcal{M}}{\Exp{q(\bz|\bx)}{ \log p(\bx|\bz) }} \\
&- \sum_{j=1}^K \text{maximum}(\lambda, \Exp{\bx \sim \mathcal{M}}{ D_{KL}(q(z_j|\bx)||p(z_j)) }
\label{eq:freebits}
\end{align}
Since increasing the latent information is generally advantageous for the first (unaffected) term of the objective (often called the \emph{negative reconstruction error}), this results in $\Exp{\bx \sim \mathcal{M}}{D_{KL}(q(\bz_j|\bx)||p(\bz_j))} \geq \lambda$ for all $j$, in practice. In ~\cite{kingma2016improving} it was found that the method worked well for a fairly wide range of values ($\lambda \in [0.125,0.25,0.5,1,2]$), resulting in significant improvement in the resulting log-likelihood on a benchmark result.

\subsection{Blurriness of generative model}
In section~\ref{sec:elbokl} we saw that optimizing the ELBO is equivalent to minimizing $D_{KL}(\allowbreak\qDP(\bx,\bz)||\pT(\bx,\bz))$. If a perfect fit between $\qDP(\bx,\bz)$ and $\pT(\bx,\bz)$ is not possible, then the variance of $\pT(\bx,\bz)$ and $\pT(\bx)$ will end up larger than the variance $\qDP(\bx,\bz)$ and the data $\qDP(\bx)$. This is due to the direction of the KL divergence; if there are values of $(\bx,\bz)$ which are likely under $\qDP$ but not under $\pT$, the term $\Exp{\qDP(\bx,\bz)}{\log \pT(\bx,\bz)}$ will go to infinity. However, the reverse is not true: the generative model is only slightly penalized when putting probability mass on values of $(\bx,\bz)$ with no support under $\qDP$.

Issues with 'blurriness' can thus can be countered by choosing a sufficiently flexible inference model, and/or a sufficiently flexible generative model. In the next two chapters we will discuss techniques for constructing flexible inference models and flexible generative models.

\section{Related prior and concurrent work}

Here we briefly discuss relevant literature prior to and concurrent with the work in ~\citep{kingma2013auto}.

The wake-sleep algorithm~\citep{hinton1995wake} is another on-line learning method, applicable to the same general class of continuous latent variable models. Like our method, the wake-sleep algorithm employs a recognition model that approximates the true posterior. A drawback of the wake-sleep algorithm is that it requires a concurrent optimization of two objective functions, which together do not correspond to optimization of (a bound of) the marginal likelihood.
An advantage of wake-sleep is that it also applies to models with discrete latent variables. Wake-Sleep has the same computational complexity as AEVB per datapoint.

Variational inference has a long history in the field of machine learning. We refer to ~\citep{wainwright2008graphical} for a comprehensive overview and synthesis of ideas around variational inference for exponential family graphical models. Among other connections, ~\citep{wainwright2008graphical} shows how various inference algorithms (such as expectation propagation, sum-product, max-product and many others) can be understood as exact or approximate forms of variational inference.

Stochastic variational inference~\cite{hoffman2013stochastic} has received increasing interest. \cite{blei2012variational} introduced a control variate schemes to reduce the variance of the score function gradient estimator, and applied the estimator to exponential family approximations of the posterior. In~\citep{ranganath2013black} some general methods, e.g. a control variate scheme, were introduced for reducing the variance of the original gradient estimator. In~\citep{salimans2013fixedform}, a similar reparameterization as in this work was used in an efficient version of a stochastic variational inference algorithm for learning the natural parameters of exponential-family approximating distributions.

In ~\cite{graves2011practical} a similar estimator of the gradient is introduced; however the estimator of the variance is not an unbiased estimator w.r.t. the ELBO gradient.

The VAE training algorithm exposes a connection between directed probabilistic models (trained with a variational objective) and autoencoders. A connection between \emph{linear} autoencoders and a certain class of generative linear-Gaussian models has long been known. In ~\citep{roweis1998algorithms} it was shown that PCA corresponds to the maximum-likelihood (ML) solution of a special case of the linear-Gaussian model with a prior $p(\bz) = \mathcal{N}(0,\bI)$ and a conditional distribution $p(\bx|\bz) = \mathcal{N}(\bx; \bW \bz, \epsilon \bI)$, specifically the case with infinitesimally small $\epsilon$.
In this limiting case, the posterior over the latent variables $p(\bz|\bx)$ is a Dirac delta distribution: $p(\bz|\bx) = \delta(\bz-\bW'\bx)$ where $\bW' = (\bW^T \bW)^{-1} \bW^T$, i.e., given $\bW$ and $\bx$ there is no uncertainty about latent variable $\bz$. ~\cite{roweis1998algorithms} then introduces an EM-type approach to learning $\bW$.
Much earlier work~\citep{bourlard1988auto} showed that optimization of linear autoencoders retrieves the principal components of data, from which it follows that learning linear autoencoders correspond to a specific method for learning the above case of linear-Gaussian probabilistic model of the data. However, this approach using linear autoencoders is limited to linear-Gaussian models, while our approach applies to a much broader class of continuous latent variable models.

When using neural networks for both the inference model and the generative model, the combination forms a type of autoencoder~\citep{goodfellow2016deeplearning} with a specific regularization term:
\begin{align}
\TELBO(\bx;\beps) =  \underbrace{ \log \pT(\bx|\bz) }_{\text{Negative reconstruction error}}
+ \underbrace{ \log \pT(\bz) - \log \qP(\bz|\bx) }_{\text{Regularization terms}}
\end{align}

\looseness=1 In an analysis of plain autoencoders~\citep{vincent2010stacked} it was shown that the training criterion of unregularized autoencoders corresponds to maximization of a lower bound (see the infomax principle~\citep{linsker1989application}) of the mutual information between input $X$ and latent representation $Z$. Maximizing (w.r.t. parameters) of the mutual information is equivalent to maximizing the conditional entropy, which is lower bounded by the expected log-likelihood of the data under the autoencoding model~\citep{vincent2010stacked}, i.e. the negative reconstruction error.
However, it is well known that this reconstruction criterion is in itself not sufficient for learning useful representations~\citep{bengio2013representation}.
Regularization techniques have been proposed to make autoencoders learn useful representations, such as  denoising, contractive and sparse autoencoder variants~ \citep{bengio2013representation}. The VAE objective contains a regularization term dictated by the variational bound, lacking the usual nuisance regularization hyper-parameter required to learn useful representations.
Related are also encoder-decoder architectures such as the predictive sparse decomposition (PSD)~\citep{koray-psd-08}, from which we drew some inspiration. Also relevant are the recently introduced Generative Stochastic Networks~\citep{bengio2014deep} where noisy autoencoders learn the transition operator of a Markov chain that samples from the data distribution. In~\citep{salakhutdinov2010efficient} a recognition model was employed for efficient learning with Deep Boltzmann Machines.
These methods are targeted at either unnormalized models (i.e. undirected models like Boltzmann machines) or limited to sparse coding models, in contrast to our proposed algorithm for learning a general class of directed probabilistic models.

\looseness=1 The proposed DARN method ~\citep{gregor2014deep}, also learns a directed probabilistic model using an autoencoding structure, however their method applies to binary latent variables. In concurrent work, ~\cite{rezende2014stochastic} also make the connection between autoencoders, directed probabilistic models and stochastic variational inference using the reparameterization trick we describe in ~\citep{kingma2013auto}. Their work was developed independently of ours and provides an additional perspective on the VAE.

\subsection{Score function estimator}
\label{sec:reinforce}

An alternative unbiased stochastic gradient estimator of the ELBO is the \emph{score function estimator} \citep{kleijnen1996optimization}:
\begin{align}
\nabla_{\bphi} \Exp{\qP(\bz|\bx)}{f(\bz)}
&= \Exp{\qP(\bz|\bx)}{f(\bz) \nabla_{\bphi} \log \qP(\bz|\bx) } \\
&\simeq f(\bz) \nabla_{\bphi} \log \qP(\bz|\bx)
\end{align}
where $\bz \sim \qP(\bz|\bx)$.

This is also known as the \emph{likelihood ratio} estimator \citep{glynn1990likelihood,fu2006gradient} and the REINFORCE gradient estimator ~\citep{williams1992simple}. The method has been successfully used in various methods like \emph{neural variational inference} \citep{mnih2014neural}, \emph{black-box variational inference} \citep{ranganath2013black}, \emph{automated variational inference}  \citep{wingate2013automated}, and \emph{variational stochastic search} \citep{paisley2012variational}, often in combination with various novel control variate techniques~\citep{glasserman2013monte} for variance reduction. An advantage of the likelihood ratio estimator is its applicability to discrete latent variables.

We do not directly compare to these techniques, since we concern ourselves with continuous latent variables, in which case we have (computationally cheap) access to gradient information $\nabla_{\bz} \log \pT(\bx,\bz)$, courtesy of the backpropagation algorithm. The score function estimator solely uses the scalar-valued $\log \pT(\bx,\bz)$, ignoring the gradient information about the function $\log \pT(\bx,\bz)$, generally leading to much higher variance. This has been experimentally confirmed by e.g.~\citep{kucukelbir2016automatic}, which finds that a sophisticated score function estimator requires two orders of magnitude more samples to arrive at the same variance as a reparameterization based estimator.

The difference in efficiency of our proposed reparameterization-based gradient estimator, compared to score function estimators, can intuitively be understood as removing an information bottleneck during the computation of gradients of the ELBO w.r.t. $\bphi$ from current parameters $\bT$: in the latter case, this computation is bottlenecked by the scalar value $\log \pT(\bx,\bz)$, while in the former case it is bottlenecked by the much wider vector $\nabla_\bz \log \pT(\bx,\bz)$.

\chapter{Beyond Gaussian Posteriors}
\label{chap:advanced_q}

In this chapter we discuss techniques for improving the flexibility of the inference model $\qP(\bz|\bx)$. Increasing the flexibility and accuracy of the inference model wel generally improve the tightness of the variational bound (ELBO), bringing it closer the true marginal likelihood objective.

\section{Requirements for Computational Tractability}
\label{sec:comp_trac}
Requirements for the inference model, in order to be able to efficiently optimize the ELBO, are that it is (1) computationally efficient to compute and differentiate its probability density $\qP(\bz|\bx)$, and (2) computationally efficient to sample from, since both these operations need to be performed for each datapoint in a minibatch at every iteration of optimization. If $\bz$ is high-dimensional and we want to make efficient use of parallel computational resources like GPUs, then parallelizability of these operations across dimensions of $\bz$ is a large factor towards efficiency. This requirement restricts the class of approximate posteriors $q(\bz|\bx)$ that are practical to use. In practice this often leads to the use of simple Gaussian posteriors. However, as explained, we also need the density $q(\bz|\bx)$ to be sufficiently flexible to match the true posterior $p(\bz|\bx)$, in order to arrive at a tight bound.

\section{Improving the Flexibility of Inference Models}

Here we will review two general techniques for improving the flexibility of approximate posteriors in the context of gradient-based variational inference: auxiliary latent variables, and normalizing flows.

\subsection{Auxiliary Latent Variables}\label{sec:auxiliarylatents}

One method for improving the flexibility of inference models, is through the introduction of \emph{auxiliary latent variables}, as explored by ~\cite{salimans2015markov},~\citep{ranganath2016hierarchical} and ~\cite{maaloe2016auxiliary}.

The methods work by augmenting both the inference model and the generative model with a continuous auxiliary variable, here denoted with $\bu$.

The inference model defines a distribution over both $\bu$ and and $\bz$, which can, for example, factorize as:
\begin{align}
\qP(\bu,\bz|\bx) = \qP(\bu|\bx)\qP(\bz|\bu,\bx)
\end{align}
This inference model augmented with $\bu$, implicitly defines a potentially powerful implicit marginal distribution:
\begin{align}
\qP(\bz|\bx) &= \int \qP(\bu,\bz|\bx) \,d\bu
\end{align}

Likewise, we introduce an additional distribution in the generative model: such that our generative model is now over the joint distribution $\pT(\bx,\bz,\bu)$. This can, for example, factorize as:
\begin{align}
\pT(\bx,\bz,\bu) = \pT(\bu|\bx,\bz)\pT(\bx,\bz)
\end{align}

The ELBO objective with auxiliary variables, given empirical distribution $\qD(\bx)$, is then (again) equivalent to minimization of a KL divergence:
\begin{align}
&\Exp{\qD(\bx)}{ \Exp{\qP(\bu,\bz|\bx)}{\log \pT(\bx,\bz,\bu) - \log \qP(\bu,\bz|\bx) } }\\
&= D_{KL}(\qDP(\bx,\bz,\bu) || \pT(\bx,\bz,\bu)
\label{eq:augmentedelbo}
\end{align}

Recall that maximization of the original ELBO objective, without auxiliary variables, is equivalent to minimization of $D_{KL}(\qDP(\bx,\bz) ||\allowbreak \pT(\bx,\bz))$, and that maximization of the expected marginal likelihood is equivalent to minimization of $D_{KL}(\qDP(\bx) || \pT(\bx))$.

We can gain additional understanding into the relationship between the objectives, through the following equation:
\begin{align}
&D_{KL}(\qDP(\bx,\bz,\bu) || \pT(\bx,\bz,\bu))\\
&\tag*{(= \text{ELBO loss with auxiliary variables})}\\
&= D_{KL}(\qDP(\bx,\bz) || \pT(\bx,\bz)) + \Exp{\qD(\bx,\bz)}{D_{KL}(\qDP(\bu|\bx,\bz) || \pT(\bu|\bx,\bz))} \nonumber\\
&\geq D_{KL}(\qDP(\bx,\bz) || \pT(\bx,\bz))\\
&\tag*{(= \text{original ELBO objective)})}\nonumber\\
&= D_{KL}(\qD(\bx) || \pT(\bx)) + \Exp{\qD(\bx)}{D_{KL}(\qDP(\bz|\bx) || \pT(\bz|\bx))}\\
&\geq D_{KL}(\qD(\bx) || \pT(\bx))\\
&\tag*{(= \text{Marginal log-likelihood objective})}\nonumber
\end{align}

From this equation it can be seen that in principle, the ELBO gets worse by augmenting the VAE with an auxiliary variable $\bu$:
\begin{align*}
D_{KL}(\qDP(\bx,\bz,\bu) || \pT(\bx,\bz,\bu)) \geq D_{KL}(\qDP(\bx,\bz) || \pT(\bx,\bz))
\end{align*}
But because we now have access to a much more flexible class of inference distributions, $\qP(\bz|\bx)$,
the original ELBO objective $D_{KL}(\qDP(\bx,\bz) ||\allowbreak \pT(\bx,\bz))$ can improve, potentially outweighing the additional cost of \\$\Exp{\qD(\bx,\bz)}{D_{KL}(\qDP(\bu|\bx,\bz) || \pT(\bu|\bx,\bz))}$. In \citep{salimans2015markov}, \citep{ranganath2016hierarchical} and \citep{maaloe2016auxiliary} it was shown that auxiliary variables can indeed lead to significant improvements in models.

The introduction of auxiliary latent variables in the graph, are a special case of VAEs with multiple layers of latent variables, which are discussed in chapter~\ref{chap:advanced_p}. In our experiment with CIFAR-10, we make use of multiple layers of stochastic variables.

\subsection{Normalizing Flows}
An alternative approach towards flexible approximate posteriors is \emph{Normalizing Flow} (NF), introduced by~\citep{rezende2015variational} in the context of stochastic gradient variational inference. In normalizing flows, we build flexible posterior distributions through an iterative procedure. The general idea is to start off with an initial random variable with a relatively simple distribution with a known (and computationally cheap) probability density function, and then apply a chain of invertible parameterized transformations $\mathbf{f}_t$, such that the last iterate $\bz_T$ has a more flexible distribution\footnote{where $\bx$ is the context, such as the value of the datapoint. In case of models with multiple levels of latent variables, the context also includes the value of the previously sampled latent variables.}:
\begin{align}
&\beps_0 \sim p(\beps)\\
&\text{for\;} t = 1 ... T:\\
&\quad\; \beps_t = \mathbf{f}_t(\beps_{t-1},\bx)\\
&\bz = \beps_T
\end{align}
The Jacobian of the transformation factorizes:
\begin{align}
\frac{d\bz}{d\beps_{0}} = \prod_{t=1}^T \frac{d\beps_t}{d\beps_{t-1}}
\end{align}
So its determinant also factorizes as well:
\begin{align}
\log \left|\det\left(\frac{d\bz}{d\beps_{0}}\right)\right| = \sum_{t=1}^T \log \left|\det\left(\frac{d\beps_t}{d\beps_{t-1}}\right)\right|
\end{align}
As long as the Jacobian determinant of each of the transformations $\mathbf{f}_t$ can be computed, we can still compute the probability density function of the last iterate:
\begin{align}
\log \qP(\bz|\bx) = \log p(\beps_0) - \sum_{t=1}^T \log \det \left| \frac{d\beps_t}{d\beps_{t-1}} \right|
\label{eq:nf_it}
\end{align}

\cite{rezende2015variational} experimented with a transformation of the form:
\begin{align}
\mathbf{f}_t(\beps_{t-1}) = \beps_{t-1} + \mathbf{u}h(\bw^T\beps_{t-1}+b)
\label{eq:planarflow}
\end{align}
where $\mathbf{u}$ and $\bw$ are vectors, $\bw^T$ is $\bw$ transposed, $b$ is a scalar and $h(.)$ is a nonlinearity, such that $\mathbf{u}h(\bw^T\bz_{t-1}+b)$ can be interpreted as a MLP with a bottleneck hidden layer with a single unit. This flow does not scale well to a high-dimensional latent space: since information goes through the single bottleneck, a long chain of transformations is required to capture high-dimensional dependencies. 

\section{Inverse Autoregressive Transformations}
\label{sec:white}

In order to find a type of normalizing flow that scales well to a high-dimensional space, ~\cite{kingma2016improving} consider Gaussian versions of autoregressive autoencoders such as MADE ~\citep{germain2015made} and the PixelCNN~\citep{pixelrnn}. Let $\by$ be a variable modeled by such a model, with some chosen ordering on its elements $\by = \{y_i\}_{i=1}^D$. We will use $[\bmu(\by), \bsigma(\by)]$ to denote the function of the vector $\by$, to the vectors $\bmu$ and $\bsigma$. Due to the autoregressive structure, the Jacobian matrix is triangular with zeros on the diagonal: $\partial [\bmu_i, \bsigma_i] / \partial \by_j = [0,0]$ for $j \geq i$. The elements $[\mu_i(\by_{1:i-1}), \sigma_i(\by_{1:i-1})]$ are the predicted mean and standard deviation of the $i$-th element of $\by$, which are functions of only the previous elements in $\by$.

Sampling from such a model is a sequential transformation from a noise vector $\bepsilon \sim \mathcal{N}(0,\mathbf{I})$ to the corresponding vector $\by$: $y_{0} = \mu_0 + \sigma_{0} \odot \epsilon_{0}$, and for $i>0$, $y_i = \mu_i(\by_{1:i-1}) + \sigma_i(\by_{1:i-1}) \cdot \epsilon_i$. The computation involved in this transformation is clearly proportional to the dimensionality $D$. Since variational inference requires sampling from the posterior, such models are not interesting for direct use in such applications. However, the inverse transformation is interesting for normalizing flows. As long as we have $\sigma_i > 0$ for all $i$, the sampling transformation above is a one-to-one transformation, and can be inverted:
\begin{align}
\epsilon_i = \frac{y_i - \mu_i(\by_{1:i-1})}{\sigma_i(\by_{1:i-1})}
\end{align}

~\cite{kingma2016improving} make two key observations, important for normalizing flows. The first is that this inverse transformation can be parallelized, since (in case of autoregressive autoencoders) computations of the individual elements $\epsilon_i$ do not depend on each other. The vectorized transformation is:
\begin{align}
\bepsilon = (\by - \bmu(\by))/\bsigma(\by)
\label{eq:whitening}
\end{align}
where the subtraction and division are element-wise.

The second key observation, is that this inverse autoregressive operation has a simple Jacobian determinant. Note that due to the autoregressive structure, $\partial [\mu_i, \sigma_i] / \partial y_j = [0,0]$ for $j \geq i$. As a result, the transformation has a lower triangular Jacobian ($\partial \epsilon_i / \partial y_j = 0$ for $j>i$), with a simple diagonal: $\partial \epsilon_i / \partial y_i = \frac{1}{\sigma_i}$. The determinant of a lower triangular matrix equals the product of the diagonal terms. As a result, the log-determinant of the Jacobian of the transformation is remarkably simple and straightforward to compute:
\begin{align}
\log \det \left| \frac{d\bepsilon}{d\by} \right| = \sum_{i=1}^D - \log \sigma_{i}(\by)
\end{align}
The combination of model flexibility, parallelizability across dimensions, and simple log-determinant, makes this transformation interesting for use as a normalizing flow over high-dimensional latent space.

For the following section we will use a slightly different, but equivalently flexible, transformation of the type:
\begin{align}
\bepsilon = \bsigma(\by) \odot \by + \bmu(\by)
\label{eq:whitening2}
\end{align}
With corresponding log-determinant:
\begin{align}
\log \det \left| \frac{d\bepsilon}{d\by} \right| = \sum_{i=1}^D \log \sigma_{i}(\by)
\end{align}

\section{Inverse Autoregressive Flow (IAF)}

\begin{algorithm}[t]
	\SetKwFunction{EncoderNN}{EncoderNN}
	\SetKwFunction{AutoregressiveNN}{AutoregressiveNN}
	\SetKwFunction{sigmoid}{sigmoid}
	\SetKwFunction{dosum}{sum}
	\SetKwFunction{sigmoid}{sigmoid}
	\SetKwInOut{Input}{input}
	\SetKwInOut{Output}{output}
	\KwIn{$\EncoderNN(\bx;\bT)$ is an encoder neural network, with additional output $\bh$.}
	\KwIn{$\AutoregressiveNN(\bz; \bh, t, \bT)$ is a neural network that is autoregressive over $\bz$, with additional inputs $\bh$ and $t$.}
	\KwIn{$T$ signifies the number of steps of flow.}
	\KwData{\\
		\hspace{5mm}$\bx$: a datapoint, and optionally other conditioning information\\
		\hspace{5mm}$\bT$: neural network parameters
	}
	\KwResult{\\
		\hspace{5mm}$\bz$: a random sample from $q(\bz|\bx)$, the approximate posterior distribution\\
		\hspace{5mm}$l$: the scalar value of $\log q(\bz|\bx)$, evaluated at sample '$\bz$'
	}
	\BlankLine
	$[\bmu, \bsigma, \bh] \leftarrow \EncoderNN(\bx;\bT)$\\
	$\bepsilon \sim \mathcal{N}(0, I)$\\
	$\bz \leftarrow \bsigma \odot \bepsilon + \bmu$\\
	$l \leftarrow - \sum_i(\log \sigma_i + \tfrac{1}{2}\epsilon_i^2 + \tfrac{1}{2}\log(2 \pi))$\\
	\For{$t \leftarrow 1$ \KwTo $T$}{
		$[\bmm, \bs] \leftarrow \AutoregressiveNN(\bz; \bh, t, \bT)$\\
		$\bsigma \leftarrow (1 + \exp(-\bs))^{-1}$\\
		$\bz \leftarrow \bsigma \odot \bz + (1 - \bsigma) \odot \bmm$\\
		$l \leftarrow l - \sum_i(\log \sigma_i)$
	}
	\caption{
	Pseudo-code of an approximate posterior with Inverse Autoregressive Flow (IAF).
	}
	\label{algorithm:iaf}
\end{algorithm}

\begin{figure}[t]
	\centering
	\includegraphics[width=.99\textwidth]{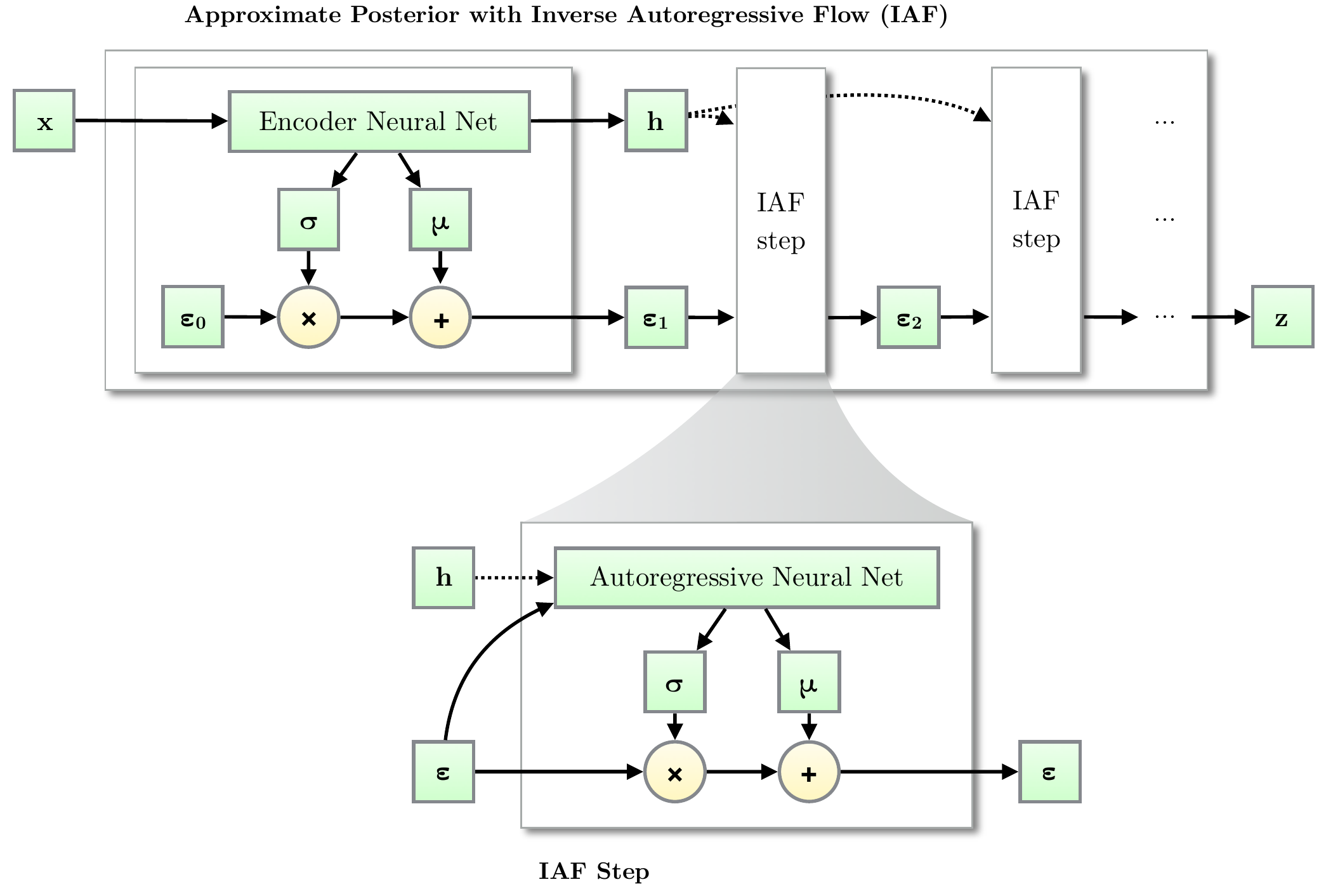}
	\caption{Like other normalizing flows, drawing samples from an approximate posterior with Inverse Autoregressive Flow (IAF) \citep{kingma2016improving} starts with a distribution with tractable density, such as a Gaussian with diagonal covariance, followed by a chain of nonlinear invertible transformations of $\bz$, each with a simple Jacobian determinant. The final iterate has a flexible distribution.}
	\label{fig:iaf_step}
\end{figure}

~\cite{kingma2016improving} propose inverse autoregressive flow (IAF) based on a chain of transformations that are each equivalent to an inverse autoregressive transformation of eq.~\eqref{eq:whitening} and eq.~\eqref{eq:whitening2}. See algorithm ~\ref{algorithm:iaf} for pseudo-code of an approximate posterior with the proposed flow. We let an initial encoder neural network output $\bmu_0$ and $\bsigma_0$, in addition to an extra output $\bh$, which serves as an additional input to each subsequent step in the flow. The chain is initialized with a factorized Gaussian $\qP(\bz_0|\bx) = \mathcal{N}(\bmu_0, \text{diag}(\bsigma_0)^2)$:
\begin{align}
\beps_0 &\sim \mathcal{N}(0,I)\\
(\bmu_0, \log \bsigma_0, \bh) &= \text{EncoderNeuralNet}(\bx; \bT)\\
\bz_0 &= \bmu_0 + \bsigma_0 \odot \bepsilon_0
\label{eq:transform}
\end{align}
IAF then consists of a chain of $T$ of the following transformations:
\begin{align}
(\bmu_t, \bsigma_t) &= \text{AutoregressiveNeuralNet}_t(\beps_{t-1}, \bh; \bT)\\
\beps_t &= \bmu_t + \bsigma_t \odot \beps_{t-1}
\label{eq:iaf}
\end{align}
\looseness=-1 Each step of this flow is an inverse autoregressive transformation of the type of eq. \eqref{eq:whitening} and eq. \eqref{eq:whitening2}, and each step uses a separate autoregressive neural network. Following eq.~\eqref{eq:nf_it}, the density under the final iterate is:
\begin{align}
\bz &\equiv \beps_T\\
\log q(\bz|\bx) &= - \sum_{i=1}^D \left(\tfrac{1}{2}\epsilon_i^2 + \tfrac{1}{2}\log(2 \pi) + \sum_{t=0}^T \log \sigma_{t,i} \right)
\label{eq:iaf_it}
\end{align}

The flexibility of the distribution of the final iterate $\beps_T$, and its ability to closely fit to the true posterior, increases with the expressivity of the autoregressive models and the depth of the chain. See figure~\ref{fig:iaf_step} for an illustration of the computation.

A numerically stable version, inspired by the LSTM-type update, is where we let the autoregressive network output $(\bmm_t,\bs_t)$, two unconstrained real-valued vectors, and compute $\beps_t$ as:
\begin{align}
(\bmm_t,\bs_t) &= \text{AutoregressiveNeuralNet}_t(\beps_{t-1}, \bh; \bT)\\
\bsigma_t &= \text{sigmoid}(\bs_t)\\
\beps_t &= \bsigma_t \odot \beps_{t-1} + (1 - \bsigma_t) \odot \bmm_t
\end{align}
This version is shown in algorithm~\ref{algorithm:iaf}. Note that this is just a particular version of the update of eq.~\eqref{eq:iaf}, so the simple computation of the final log-density of eq.~\eqref{eq:iaf_it} still applies.

It was found beneficial for results to parameterize or initialize the parameters of each $\text{AutoregressiveNeuralNet}_t$ such that its outputs $\bs_t$ are, before optimization, sufficiently positive, such as close to +1 or +2. This leads to an initial behavior that updates $\beps$ only slightly with each step of IAF. Such a parameterization is known as a 'forget gate bias' in LSTMs, as investigated by~\cite{jozefowicz2015empirical}.

\looseness=1 It is straightforward to see that a special case of IAF with one step, and a linear autoregressive model, is the fully Gaussian posterior discussed earlier. This transforms a Gaussian variable with diagonal covariance, to one with linear dependencies, i.e. a Gaussian distribution with full covariance.

\looseness=1 Autoregressive neural networks form a rich family of nonlinear transformations for IAF. For non-convolutional models, the family of masked autoregressive network introduced in \citep{germain2015made} was used as the autoregressive neural networks. For CIFAR-10 experiments, which benefits more from scaling to high dimensional latent space, the family of convolutional autoregressive autoencoders introduced by ~\citep{pixelrnn,van2016conditional} was used.

It was found that results improved when reversing the ordering of the variables after each step in the IAF chain. This is a volume-preserving transformation, so the simple form of eq.~\eqref{eq:iaf_it} remains unchanged.

\section{Related work}

As we explained, inverse autoregressive flow (IAF) is a member of the family of normalizing flows, first discussed in~\citep{rezende2015variational} in the context of stochastic variational inference. In \citep{rezende2015variational} two specific types of flows are introduced: planar flow (eq.~\eqref{eq:planarflow}) and radial flow. These flows are shown to be effective to problems with a relatively low-dimensional latent space. It is not clear, however, how to scale such flows to much higher-dimensional latent spaces, such as latent spaces of generative models of larger images, and how planar and radial flows can leverage the topology of latent space, as is possible with IAF. Volume-conserving neural architectures were first presented in in~\citep{deco1995higher}, as a form of nonlinear independent component analysis.

Another type of normalizing flow, introduced by \citep{dinh2014nice} (\emph{NICE}), uses similar transformations as IAF. In contrast with IAF, NICE was directly applied to the observed variables in a generative model. NICE is type of transformations that updates only half of the variables $\bz_{1:D/2}$ per step, adding a vector $f(\bz_{D/2+1:D})$ which is a neural network based function of the remaining latent variables $\bz_{D/2+1:D}$. Such large blocks have the advantage of computationally cheap inverse transformation, and the disadvantage of typically requiring longer chains. In experiments, \citep{rezende2015variational} found that this type of transformation is generally less powerful than other types of normalizing flow, in experiments with a low-dimensional latent space. Concurrently to our work, NICE was extended to high-dimensional spaces in \citep{dinh2016density} (\emph{Real NVP}).

A potentially powerful transformation is the \textit{Hamiltonian flow} used in Hamiltonian Variational Inference~\citep{salimans2015markov}. Here, a transformation is generated by simulating the flow of a Hamiltonian system consisting of the latent variables $\bz$, and a set of auxiliary momentum variables. This type of transformation has the additional benefit that it is guided by the exact posterior distribution, and that it leaves this distribution invariant for small step sizes. Such a transformation could thus take us arbitrarily close to the exact posterior distribution if we can apply it a sufficient number of times. In practice, however, Hamiltonian Variational Inference is very demanding computationally. Also, it requires an auxiliary variational bound to account for the auxiliary variables, which can impede progress if the bound is not sufficiently tight.

An alternative method for increasing the flexibility of variational inference is the introduction of auxiliary latent variables
~\citep{salimans2015markov,ranganath2016hierarchical,tran2015variational}, discussed in ~\ref{sec:auxiliarylatents}, and corresponding auxiliary inference models. Latent variable models with multiple layers of stochastic variables, such as the one used in our experiments, are often equivalent to such auxiliary-variable methods. We combine deep latent variable models with IAF in our experiments, benefiting from both techniques.

\chapter{Deeper Generative Models}
\label{chap:advanced_p}

In the previous chapter we explain advanced strategies for improving inference models. In this chapter, we review strategies for learning deeper generative models, such as inference and learning with multiple latent variables or observed variables, and techniques for improving the flexibility of the generative models $\pT(\bx,\bz)$.

\section{Inference and Learning with Multiple Latent Variables}
\label{sec:multiplelatentvariables}

The generative model $\pT(\bx,\bz)$, and corresponding inference model $\qP(\bz|\bx)$ can be parameterized as any directed graph. Both $\bx$ and $\bz$ can be composed of multiple variables with some topological ordering. It may not be immediately obvious how to optimize such models in the VAE framework; it is, however, quite straightforward, as we will now explain.

\looseness=-1 Let $\bz = \{\bz_1,...,\bz_K\}$, and $\qP(\bz|\bx) = \qP(\bz_1,...,\bz_K|\bx)$ where the subscript corresponds with the topological ordering of each variable. Given a datapoint $\bx$, computation of the ELBO estimator consists of two steps:
\begin{enumerate}
	\item Sampling $\bz \sim \qP(\bz|\bx)$. In case of multiple latent variables, this means \emph{ancestral sampling} the latent variables one by one, in topological ordering defined by the inference model's directed graph. In pseudo-code, the ancestral sampling step looks like:
	\begin{align}
	&\text{for\;} i = 1 ... K:\\
	&\quad\; \bz_i \sim \qP(\bz_i | Pa(\bz_i))
	\end{align}
	where $Pa(\bz_i)$ are the parents of variable $\bz_i$ in the inference model, which may include $\bx$. In reparameterized (and differentiable) form, this is:
	\begin{align}
	&\text{for\;} i = 1 ... K:\\
	&\quad\; \beps_i \sim p(\beps_i)\\
	&\quad\; \bz_i = \bg_i(\beps_i,Pa(\bz_i),\bphi)
	\end{align}
	\item Evaluating the scalar value $(\log \pT(\bx,\bz) - \log \qP(\bz|\bx))$ at the resulting sample $\bz$ and datapoint $\bx$. This scalar is the unbiased stochastic estimate lower bound on $\log \pT(\bx)$. It is also differentiable and optimizable with SGD.
\end{enumerate}

\subsection{Choice of ordering}

\begin{figure*}[t!]
	\centering
	\begin{subfigure}[]{0.99\textwidth}
		\centering
		\includegraphics[width=.99\textwidth]{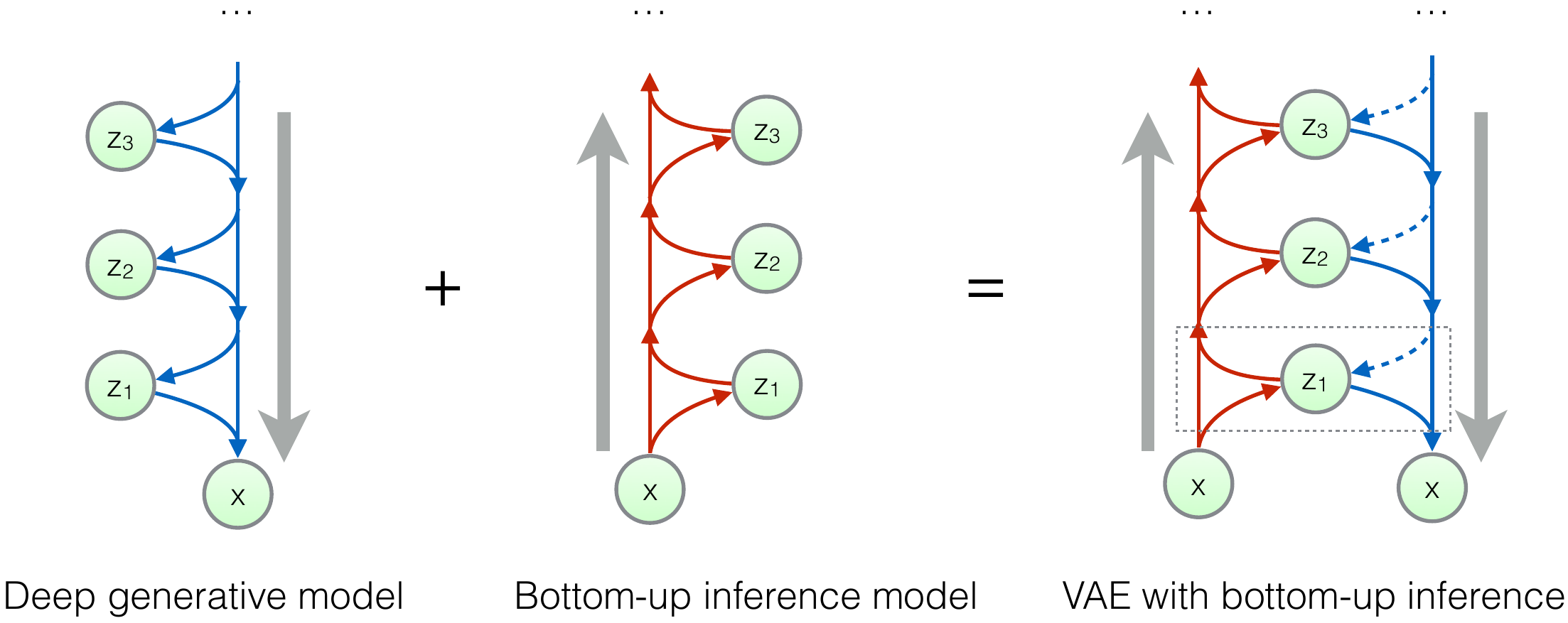}
		\caption{VAE with bottom-up inference.
		}
	\end{subfigure}%
	\vspace{5mm}
	\begin{subfigure}[]{0.99\textwidth}
		\centering
		\includegraphics[width=.99\textwidth]{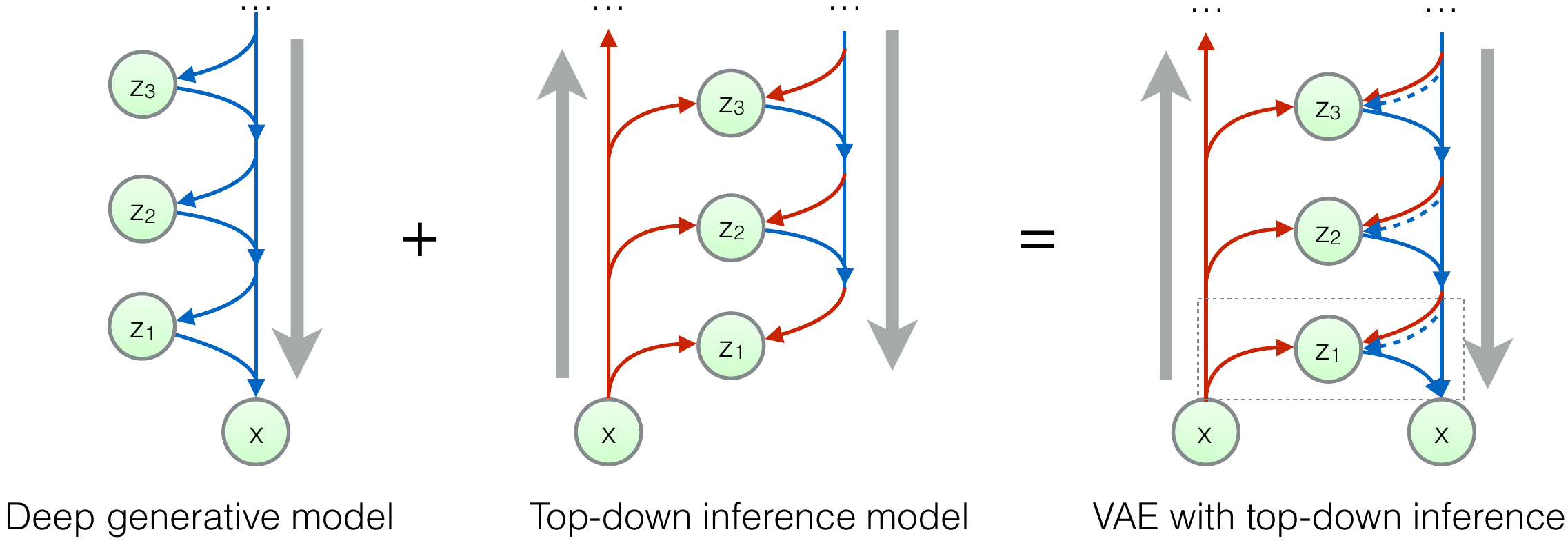}
		\caption{VAE with top-down inference.}
	\end{subfigure}%
	\vspace{5mm}
	\caption{Illustration, taken from ~\cite{kingma2016improving}, of two choices of directionality of the inference model. Sharing directionality of inference, as in (\textbf{b}), has the benefit that it allows for straightforward sharing of parameters between the generative model and the inference model.}
	\label{fig:ordering}
\end{figure*}

It should be noted that the choice of latent variables' topological ordering for the inference model can be different from the choice of ordering for the generative model. 

Since the inference model has the data as root node, while the generative model has the data as leaf node, one (in some sense) logical choice would be to let the topological ordering of the latent variables in the inference model be the reverse of the ordering in the generative model. 

In multiple works ~\citep{salimans2016structured, sonderby2016train, kingma2016improving} it has been shown that it can be advantageous to let the generative model and inference model \emph{share} the topological ordering of latent variables. The two choices of ordering are illustrated in figure ~\ref{fig:ordering}. One advantage of shared ordering, as explained in these works, is that this allows us to easily share parameters between the inference and generative models, leading to faster learning and better solutions.

To see why this might be a good idea, note that the true posterior over the latent variables, is a function of the prior:
\begin{align}
\pT(\bz|\bx) \propto \pT(\bz) \pT(\bx|\bz)
\end{align}
Likewise, the posterior of a latent variable given its parents (in the generative model), is:
\begin{align}
\pT(\bz_i|\bx,Pa(\bz_i)) \propto \pT(\bz_i|Pa(\bz_i)) \pT(\bx|\bz_i,Pa(\bz_i))
\label{eq:pseudolikelihood}
\end{align}
Optimization of the generative model changes both $\pT(\bz_i|Pa(\bz_i))$ and $\pT(\bx|\bz_i,Pa(\bz_i))$. By coupling the inference model $\qP(\bz_i|\bx,Pa(\bz_i))$ and prior $\pT(\bz_i|Pa(\bz_i))$, changes in $\pT(\bz_i|Pa(\bz_i))$ can be directly reflected in changes in $\qP(\bz_i|Pa(\bz_i))$.

This coupling is especially straightforward when $\pT(\bz_i|Pa(\bz_i))$ is Gaussian distributed. The inference model can be directly specified as the product of this Gaussian distribution, with a learned quadratic pseudo-likelihood term: 
\begin{align*}
\qP(\bz_i|Pa(\bz_i),\bx) = \pT(\bz_i|Pa(\bz_i)) \tilde{l}(\bz_i;\bx,Pa(\bz_i)) / Z ,
\end{align*}
where $Z$ is tractable to compute. This idea is explored by ~\citep{salimans2016structured} and ~\citep{sonderby2016train}. In principle this idea could be extended to a more general class of conjugate priors, but no work on this is known at the time of writing.

A less constraining variant, explored by ~\citep{kingma2016improving}, is to simply let the neural network that parameterizes $\qP(\bz_i|Pa(\bz_i),\bx)$ be partially specified by a part of the neural network that parameterizes $\pT(\bz_i|Pa(\bz_i))$. In general, we can let the two distributions share parameters. This allows for more complicated posteriors, like normalizing flows or IAF.

\section{Alternative methods for increasing expressivity}

Typically, especially with large data sets, we wish to choose an expressive class of directed models, such that it can feasibly approximate the true distribution. Popular strategies for specifying expressive models are:
\begin{itemize}
	\item Introduction of latent variables into the directed models, and optimization through (amortized) variational inference, as explained in this work.
	\item Full autoregression: factorization of distributions into univariate (one-dimensional) conditionals, or at least very low-dimensional conditionals (section~\ref{sec:armodel}). 
	\item Specification of distributions through \emph{invertible transformations with \allowbreak tractable Jacobian determinant} (section~\ref{sec:invtrans}).
\end{itemize}

Synthesis from fully autoregressive models models is relatively slow, since the \emph{length of computation} for synthesis from such models is linear in the dimensionality of the data. The length of computation of the log-likelihood of fully autoregressive models does not necesarilly scale with the dimensionality of the data. In this respect, introduction of latent variables for improving expressivity is especially interesting when $\bx$ is very high-dimensional. It is relatively straightforward and computationally attractive, due to parallelizability, to specify directed models over high-dimensional variables where each conditional factorizes into independent distributions. For example, if we let $\pT(\bx_j | Pa(\bx_j)) = \prod_k \pT(x_{j,k} | Pa(\bx_j))$, where each factor is a univariate Gaussian whose means and variance are nonlinear functions (specified by a neural network) of the parents $Pa(\bx_j)$, then computations for both synthesis and evaluation of log-likelihood can be fully parallelized across dimensions $k$. See~\citep{kingma2016improving} for experiments demonstrating a 100x improvement in speed of synthesis.

The best models to date, in terms of attained log-likelihood on test data, employ a combination of the three approaches listed above.

\section{Autoregressive Models}
\label{sec:armodel}

A powerful strategy for modeling high-dimensional data is to divide up the high-dimensional observed variables into small constituents (often single dimensional parts, or otherwise just parts with a small number of dimensions), impose a certain ordering, and to model their dependencies as a directed graphical model. The resulting directed graphical model breaks up the joint distribution into a product of a factors:
\begin{align}
    \pT(\bx) = \pT(x_1,...,x_D) = \pT(x_1) \prod_{j=2}^T \pT(x_j|Pa(\bx_j))
    \label{eq:armodel}
\end{align}
where $D$ is the dimensionality of the data. This is known as an \emph{autoregressive (AR) model}. In case of neural network based autoregressive models, we let the conditional distributions be parameterized with a neural network:
\begin{align}
    \pT(x_j|\bx_{< j}) = \pT(x_j|\NeuralNet_{\bT}^j(Pa(\bx_j)))
    \label{eq:univariate}
\end{align}
In case of continuous data, autoregressive models can be interpreted as a special case of a more general approach: learning an invertible transformation from the data to a simpler, known distribution such as a Gaussian or Uniform distribution; this approach with invertible transformations is discussed in section ~\ref{sec:invtrans}. The techniques of autoregressive models and invertible transformations can be naturally combined with variational autoencoders, and at the time of writing, the best systems use a combination~\cite{rezende2015variational, kingma2016improving, gulrajani2016pixelvae}.

A disadvantage of autoregressive models, compared to latent-variable models, is that ancestral sampling from autoregressive models is a sequential operation computation of $\mathcal{O}(D)$ length, i.e. proportional to the dimensionality of the data. Autoregressive models also require choosing a specific ordering of input elements (equation~\eqref{eq:armodel}). When no single natural one-dimensional ordering exists, like in two-dimensional images, this leads to a model with a somewhat awkward inductive bias.

\section{Invertible transformations with tractable Jacobian determinant}
\label{sec:invtrans}
In case of continuous data, autoregressive models can be interpreted as a special case of a more general approach: learning an invertible transformation with tractable Jacobian determinant (also called \emph{normalizing flow}) from the data to a simpler, known distribution such as a Gaussian or Uniform distribution.  If we use neural networks for such invertible mappings, this is a powerful and flexible approach towards probabilistic modeling of continuous data and nonlinear independent component analysis~\citep{deco1995higher}.

Such normalizing flows iteratively update a variable, which is constrained to be of the same dimensionality as the data, to a target distribution. This constraint on the dimensionality of intermediate states of the mapping can make such transformations more challenging to optimize than methods without such constraint. An obvious advantage, on the other hand, is that the likelihood and its gradient are tractable. In \citep{dinh2014nice,dinh2016density}, particularly interesting flows (\emph{NICE} and \emph{Real NVP}) were introduced, with equal computational cost and depth in both directions, making it both relatively cheap to optimize and to sample from such models. At the time of writing, no such model has yet been demonstrated to lead to the similar performance as purely autoregressive or VAE-based models in terms of data log-likelihood, but this remains an active area of research.

\section{Follow-Up Work}
\label{sec:applications}

Some important applications and motivations for deep generative models and variational autoencoders are:
\begin{itemize}
	\item Representation learning: learning better representations of the data. Some uses of this are:
	\begin{itemize}
		\item Data-efficient learning, such as semi-supervised learning
		\item Visualisation of data as low-dimensional manifolds
	\end{itemize}
	\item Artificial creativity: plausible interpolation between data and extrapolation from data.
\end{itemize}  

Here we will now highlight some concrete applications to representation learning and artificial creativity.

\subsection{Representation Learning}

In the case of \emph{supervised learning}, we typically aim to learn a conditional distribution: to predict the distribution over the possible values of a variable, given the value of some another variable. One such problem is that of image classification: given an image, the prediction of a distribution over the possible class labels. Through the yearly ImageNet competion~\citep{russakovsky2015imagenet}, it has become clear that deep convolutional neural networks~\citep{lecun1998gradient,goodfellow2016deeplearning} (CNNs), \emph{given a large amount of labeled images}, are extraordinarily good at solving the image classification task. Modern versions of CNNs based on residual networks, which is a variant of LSTM-type neural networks~\citep{hochreiter1997long}, now arguably achieves human-level classification accuracy on this task~\citep{he2015delving, he2015deep}.

When the number of labeled examples is low, solutions found with purely supervised approaches tend to exhibit poor generalization to new data. In such cases, generative models can be employed as an effective type of regularization. One particular strategy, presented in ~\cite{kingma2014semi}, is to optimize the classification model jointly with a variational autoencoder over the input variables, sharing parameters between the two. The variational autoencoder, in this case, provides an auxiliary objective, improving the data efficiency of the classification solution. Through sharing of statistical strength between modeling problems, this can greatly improve upon the supervised classification error. Techniques based on VAEs are now among state of the art for semi-supervised classification~\citep{maaloe2016auxiliary}, with on average under 1\% classification error in the MNIST classification problem, when trained with only 10 labeled images per class, i.e. when more than 99.8\% of the labels in the training set were removed. In concurrent work~\citep{rezende2016one}, it was shown that VAE-based semi-supervised learning can even do well when only a single sample per class is presented. 

A standard supervised approach, \emph{GoogLeNet}~\citep{szegedy2015going}, which normally achieves near state-of-the-art performance on the ImageNet validation set, achieves only around 5\% top-1 classification accuracy when trained with only 1\% of the labeled images, as shown by~\cite{pu2016variational}. In contrast, they show that a semi-supervised approach with VAEs achieves around 45\% classification accuracy on the same task, when modeling the labels jointly with the labeled and unlabeled input images.

\subsection{Understanding of data, and artificial creativity}

Generative models with latent spaces allow us to transform the data into a simpler latent space, explore it in that space, and understand it better. A related branch of applications of deep generative models is the synthesis of plausible pseudo-data with certain desirable properties, sometimes coined as \emph{artificial creativity}.

\subsubsection{Chemical Design}
\begin{figure}
	\centering
	\includegraphics[width=0.65\textwidth]{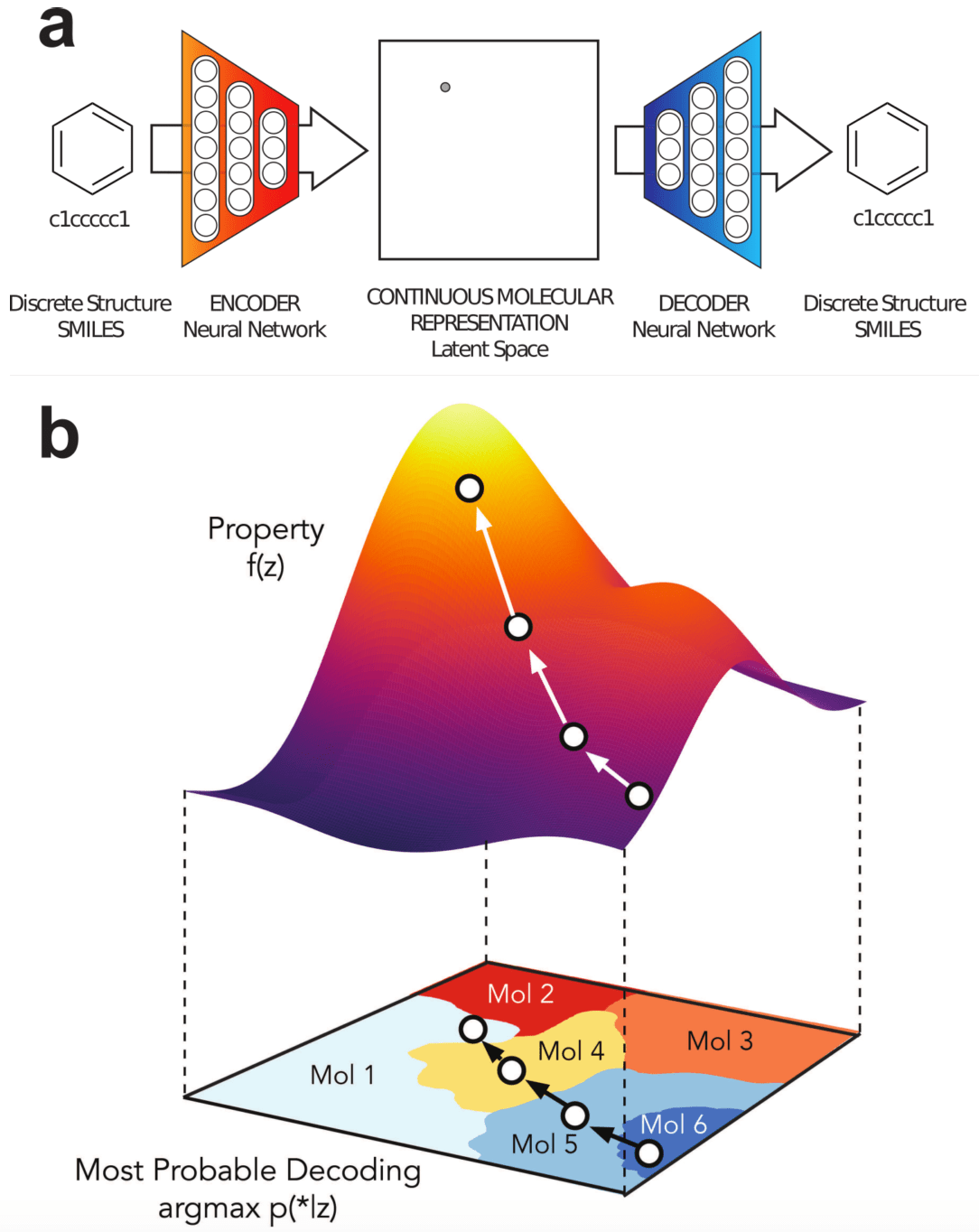}
	\caption{(a) Application of a VAE to chemical design in~\citep{gomez2016automatic}. A latent continuous representation $\bz$ of molecules is learned on a large dataset of molecules. (b) This continuous representation enables gradient-based search of new molecules that maximizes $f(\bz)$, a certain desired property.}
	\label{fig:molecules}
\end{figure}

One example of a recent scientific application of artificial creativity, is shown in ~\cite{gomez2016automatic}. In this paper, a fairly straightforward VAE is trained on hundreds of thousands of existing chemical structures. The resulting continuous representation (latent space) is subsequently used to perform gradient-based optimization towards certain properties; the method is demonstrated on the design of drug-like molecules and organic light-emitting diodes. See figure~\ref{fig:molecules}.

\subsubsection{Natural Language Synthesis}
\begin{figure}
	\centering
	\includegraphics[width=0.5\textwidth]{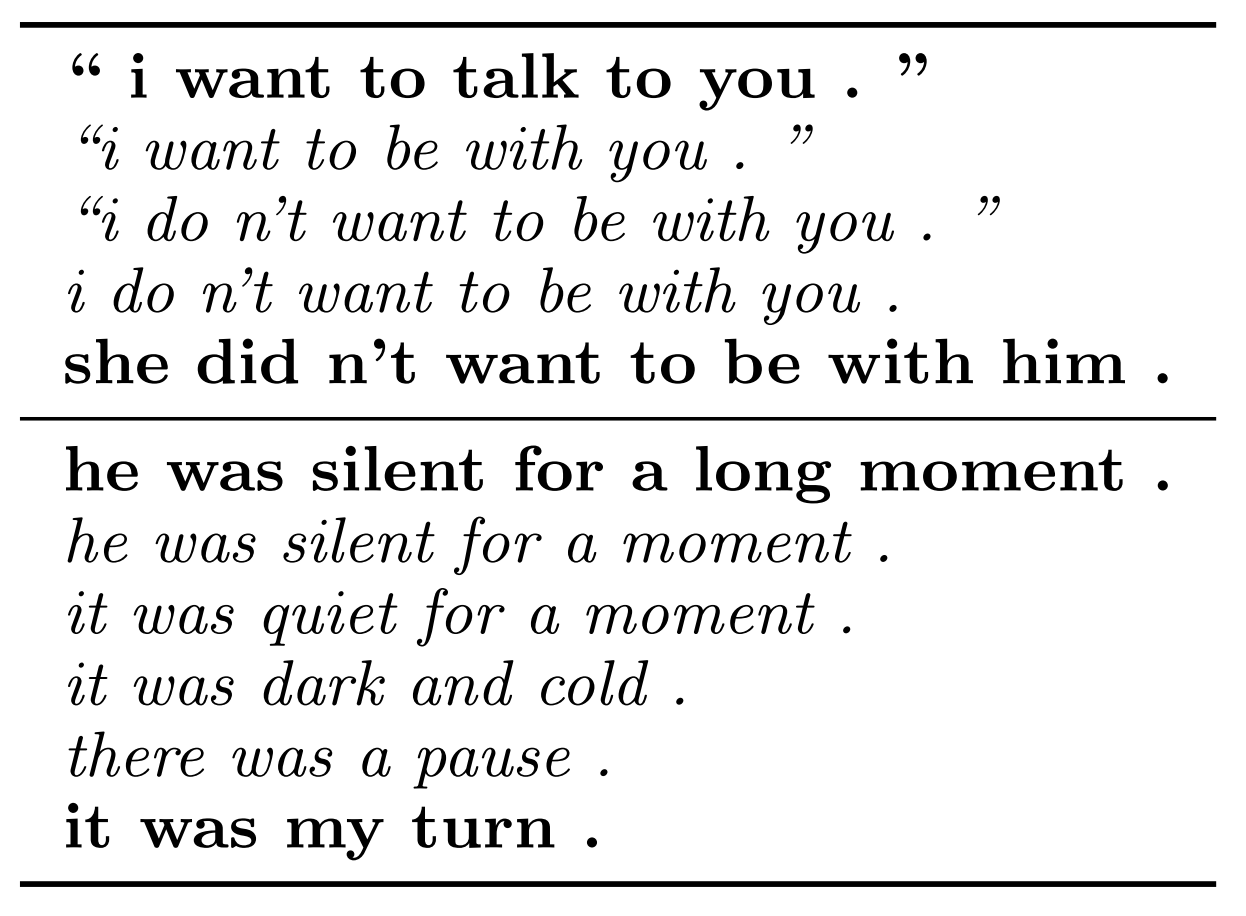}
	\caption{An application of VAEs to interpolation between pairs of sentences, from~\citep{bowman2015generating}. The intermediate sentences are grammatically correct, and the topic and syntactic structure are typically locally consistent.}
	\label{fig:sentences}
\end{figure}

A similar approach was used to generating natural-language sentences from a continuous space by~\cite{bowman2015generating}. In this paper, it is shown how a VAE can be successfully trained on text. The model is shown to succesfully interpolate between sentences, and for imputation of missing words. See figure~\ref{fig:sentences}.

\subsubsection{Astronomy}
In~\citep{ravanbakhsh2016enabling}, VAEs are applied to simulate observations of distant galaxies. This helps with the calibration of systems that need to indirectly detect the shearing of observations of distant galaxies, caused by weak gravitational lensing in the presence of dark matter between earth and those galaxies. Since the lensing effects are so weak, such systems need to be calibrated with ground-truth images with a known amount of shearing. Since real data is still limited, the proposed solution is to use deep generative models for synthesis of pseudo-data.

\subsubsection{Image (Re-)Synthesis}
\begin{figure}
	\centering
	\includegraphics[width=0.7\textwidth]{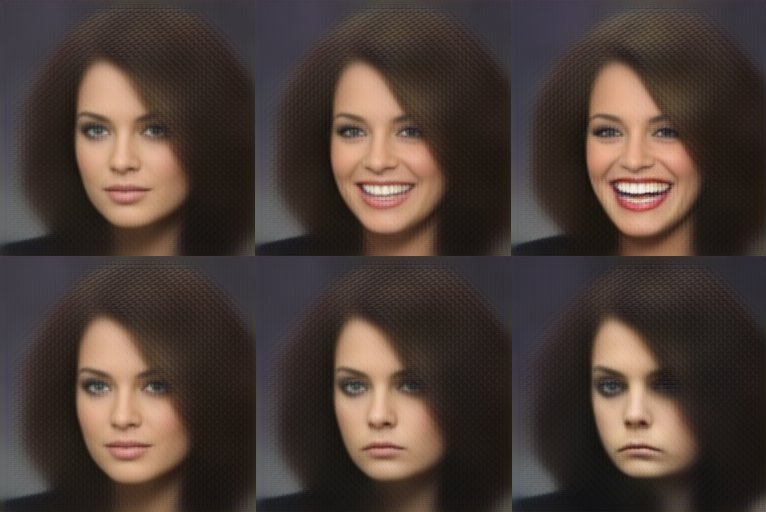}
	\caption{VAEs can be used for image resynthesis. In this example by ~\cite{white2016sampling}, an original image (left) is modified in a latent space in the direction of a \emph{smile vector}, producing a range of versions of the original, from smiling to sadness.}
	\label{fig:happysad}
\end{figure}

A popular application is image (re)synthesis. One can optimize a VAE to form a generative model over images. One can synthesize images from the generative model, but the inference model (or \emph{encoder}) also allows one to encode real images into a latent space. One can modify the encoding in this latent space, then decode the image back into the observed space. 
Relatively simple transformations in the observed space, such as linear transformations, often translate into semantically meaningful modifications of the original image. 
One example, as demonstrated by \cite{white2016sampling}, is the modification of images in latent space along a "smile vector" in order to make them more happy, or more sad looking. See figure \ref{fig:happysad} for an example. 

\subsection{Other relevant follow-up work}

We unfortunately do not have space to discuss all follow-up work in depth, but will here  highlight a selection of relevant recent work.

In addition to our original publication~\citep{kingma2013auto}, two later papers have proposed equivalent algorithms~\citep{rezende2014stochastic,lazaro2014doubly}, where the latter work applies the same reparameterization gradient method to the estimation of parameter posteriors, rather than amortized latent-variable inference.

In the appendix of~\citep{kingma2013auto} we proposed to apply the reparameterization gradients to estimation of parameter posteriors. In~\citep{blundell2015weight} this method, with a mixture-of-Gaussians prior and named \emph{Bayes by Backprop}, was used in experiments with some promising early results. In~\citep{kingma2015variational} we describe a refined method, the \emph{local reparameterization trick}, for further decreasing the variance of the gradient estimator, and applied it to estimation of Gaussian parameter posteriors. Further results were presented in~\citep{louizos2017bayesian,louizos2017multiplicative,louizos2016structured} with increasingly sophisticated choices of priors and approximate posteriors. In ~\citep{kingma2015variational,gal2016theoretically}, a similar reparameterization was used to analyze Dropout as a Bayesian method, coined \emph{Variational Dropout}. In ~\citep{molchanov2017variational} this method was further analyzed and refined. Various papers have applied reparameterization gradients for estimating parameter posteriors, including~\citep{fortunato2017bayesian} in the context of recurrent neural networks and~\citep{kucukelbir2016automatic} more generally for Bayesian models and in~\citep{tran2017deep} for deep probabilistic programming. A Bayesian nonparametric variational family based in the Gaussian Process using reparameterization gradients was proposed in~\citep{tran2015variational}.

Normalizing flows~\citep{rezende2015variational} were proposed as a framework for improving the flexibility of inference models. In ~\cite{kingma2016improving}, the first normalizing flow was proposed that scales well to high-dimensional latent spaces. The same principle was later applied in~\citep{papamakarios2017masked} for density estimation, and further refined in~\citep{huang2018neural}. Various other flows were proposed in~\citep{tomczak2016improving,tomczak2017improving} and ~\citep{berg2018sylvester}.

As an alternative to (or in conjunction with) normalizing flows, one can use auxiliary variables to improve posterior flexibility. This principle was, to the best of our knowledge, first proposed in ~\cite{salimans2015markov}. In this paper, the principle was used in a combination of variational inference with Hamiltonian Monte Carlo (HMC), with the momentum variables of HMC as auxiliary variables. Auxiliary variables were more elaborately discussed in in~\citep{maaloe2016auxiliary} as Auxiliary Deep Generative Models. Similarly, one can use deep models with multiple stochastic layers to improve the variational bound, as demonstrated in ~\citep{sonderby2016train} and ~\citep{sonderby2016ladder} as Ladder VAEs.

There has been plenty of follow-up work on gradient variance reduction for the variational parameters of discrete latent variables, as opposed to continuous latent variables for which reparameterization gradients apply. These proposals include NVIL ~\citep{mnih2014neural}, MuProp ~\citep{gu2015muprop}, Variational inference for Monte Carlo objectives~\citep{mnih2016variational}, the Concrete distribution ~\citep{maddison2016concrete} and Categorical Reparameterization with Gumbel-Softmax~\citep{jang2016categorical}.

The ELBO objective can be generalized into an importance-weighted objective, as proposed in ~\citep{burda2015importance} (Importance-Weighted Autoencoders). This potentially reduces the variance in the gradient, but has not been discussed in-depth here since (as often the case with importance-weighted estimators) it can be difficult to scale to high-dimensional latent spaces. Other objectives have been proposed such as R\'enyi divergence variational inference~\citep{li2016renyi}, Generative Moment Matching Networks~\citep{li2015generative}, objectives based on normalizing such as NICE and RealNVP flows~\citep{sohl2015deep,dinh2014nice}, black-box $\alpha$-divergence minimization~\citep{hernandez2016black} and Bi-directional Helmholtz Machines~\citep{bornschein2016bidirectional}.

Various combinations with adversarial objectives have been proposed. In ~\citep{makhzani2015adversarial}, the "adversarial autoencoder" (AAE) was proposed, a probabilistic autoencoder that uses a generative adversarial network (GAN)~\citep{goodfellow2014generative} to perform variational inference. In ~\citep{dumoulin2016adversarially} Adversarially Learned Inference (ALI) was proposed, which aims to minimize a GAN objective between the joint distributions $\qPhi(\bx,\bz)$ and $\pT(\bx,\bz)$. Other hybrids have been proposed as well~\citep{larsen2015autoencoding,brock2016neural,hsu2017voice}. 

One of the most prominent, and most difficult, applications of generative models is image modeling. In~\citep{kulkarni2015deep} (Deep convolutional inverse graphics network), a convolutional VAE was applied to modeling images with some success, building on work by ~\citep{dosovitskiy2015learning} proposing convolutional networks for image synthesis. In ~\citep{gregor2015draw} (DRAW), an attention mechanism was combined with a recurrent inference model and recurrent generative model for image synthesis. This approach was further extended in ~\citep{gregor2016towards} (Towards Conceptual Compression) with convolutional networks, scalable to larger images, and applied to image compression. In~\citep{kingma2016improving}, deep convolutional inference models and generative models were also applied to images. Furthermore, ~\citep{gulrajani2016pixelvae} (PixelVAE) and ~\citep{chen2016variational} (Variational Lossy Autoencoder) combined convolutional VAEs with the PixelCNN model~\citep{pixelrnn,van2016conditional}. Methods and VAE architectures for controlled image generation from attributes or text were studied in~\citep{kingma2014semi,yan2016attribute2image,mansimov2015generating,brock2016neural,white2016sampling}. Predicting the color of pixels based on a grayscale image is another promising application~\citep{deshpande2016learning}. The application to semi-supervised learning has been studied in ~\citep{kingma2014semi,pu2016variational,xu2017variational} among other work.

Another prominent application of VAEs is modeling of text and or sequential data~\citep{bayer2014learning,bowman2015generating, serban2016hierarchical,johnson2016composing,karl2016deep,fraccaro2016sequential, miao2016neural,semeniuta2017hybrid,zhao2017learning,yang2017improved,hu2017controllable}. VAEs have also been applied to speech and handwriting ~\cite{chung2015recurrent}. Sequential models typically use recurrent neural networks, such as LSTMs~\citep{hochreiter1997long}, as encoder and/or decoder. When modeling sequences, the validity of a sequence can sometimes be constrained by a context-free grammar. In this case, incorporation of the grammar in VAEs can lead to better models, as shown in ~\citep{kusner2017grammar} (Grammar VAEs), and applied to modeling molecules in textual representations.

Since VAEs can transform discrete observation spaces to continuous latent-variable spaces with approximately known marginals, they are interesting for use in model-based control \citep{watter2015embed,pritzel2017neural}. In ~\citep{heess2015learning} (Stochastic Value Gradients) it was shown that the re-parameterization of the observed variables, together with an observation model, can be used to compute novel forms of policy gradients. Variational inference and reparameterization gradients have also been used for variational information maximisation for intrinsically motivated reinforcement learning~\citep{mohamed2015variational} and VIME~\citep{houthooft2016vime} for improved exploration.  Variational autoencoders have also been used as components in models that perform iterative reasoning about objects in a scene~\citep{eslami2016attend}.

In~\citep{higgins2016beta} ($\beta$-VAE) it was proposed to strengthen the contribution of $D_{KL}(\qP(\bz|\bx)||\pT(\bz))$, thus restricting the information flow through the latent space, which was shown to improve disentanglement of latent factors, further studied in~\citep{chen2018isolating}.

Other applications include modeling of graphs~\citep{kipf2016variational} (Variational Graph Autoencoders), learning of 3D structure from images~\citep{rezende2016unsupervised}, one-shot learning~\citep{rezende2016one}, learning nonlinear state space models~\citep{krishnan2017structured}, voice conversion from non-parallel corpora~\citep{hsu2016voice}, discrimination-aware (fair) representations ~\citep{louizos2015variational} and transfer learning ~\citep{edwards2016towards}.

The reparameterization gradient estimator discussed in this work has been extended in various directions~\citep{ruiz2016generalized}, including acceptance-rejection sampling algorithms
~\citep{naesseth2017reparameterization}. The gradient variance can in some cases be reduced by 'carving up the ELBO' ~\citep{hoffman2016elbo,roeder2017sticking} and using a modified gradient estimator. A second-order gradient estimator has also been proposed in~\citep{fan2015fast}.

All in all, this remains an actively researched area with frequently exciting developments.

\chapter{Conclusion}
\label{chap:conclusion}

Directed probabilistic models form an important aspect of modern artificial intelligence. Such models can be made incredibly flexible by parameterizing the conditional distributions with differentiable deep neural networks.

\looseness=-1 Optimization of such models towards the maximum likelihood objective is straightforward in the fully-observed case. However, one is often more interested in flexible models with latent variables, such as deep latent-variable models, or Bayesian models with random parameters. In both cases one needs to perform approximate posterior estimation for which variational inference (VI) methods are suitable. 
In VI, inference is cast as an optimization problem over newly introduced variational parameters, typically optimized towards the ELBO, a lower bound on the model evidence, or marginal likelihood of the data.  Existing methods for such posterior inference were either relatively inefficient, or not applicable to models with neural networks as components. Our main contribution is a framework for efficient and scalable gradient-based variational posterior inference and approximate maximum likelihood learning. 

In this work we describe the variational autoencoder (VAE) and some of its extensions. A VAE is a combination of a \emph{deep latent-variable model} (DLVM) with continuous latent variables, and an associated \emph{inference model}. The DLVM is a type of generative model over the data. The inference model, also called \emph{encoder} or \emph{recognition model}, approximates the posterior distribution of the latent variables of the generative model. Both the generative model and the inference model are \emph{directed graphical models} that are wholly or partially parameterized by deep neural networks. The parameters of the models, including the parameters of the neural networks such as the weights and biases, are jointly optimized by performing stochastic gradient ascent on the so-called \emph{evidence lower bound} (ELBO). The ELBO is a lower bound on the marginal likelihood of the data, also called the \emph{variational lower bound}. Stochastic gradients, necessary for performing SGD, are obtained through a basic \emph{reparameterization trick}. The VAE framework is now a commonly used tool for various applications of probabilistic modeling and artificial creativity, and basic implementations are available in most major deep learning software libraries.

For learning flexible inference models, we proposed inverse autoregressive flows (IAF), a type of normalizing flow that allows scaling to high-dimensional latent spaces. 
An interesting direction for further exploration is comparison with transformations with computationally cheap inverses, such as NICE~\citep{dinh2014nice} and Real NVP~\citep{dinh2016density}. Application of such transformations in the VAE framework can potentially lead to relatively simple VAEs with a combination of powerful posteriors, priors and decoders. Such architectures can potentially rival or surpass purely autoregressive architectures \citep{van2016conditional}, while allowing much faster synthesis.

The proposed VAE framework remains the only framework in the literature that allows for both discrete and continuous observed variables, allows for efficient amortized latent-variable inference and fast synthesis, and which can produce close to state-of-the-art performance in terms of the log-likelihood of data.

\begin{acknowledgements}
We are grateful for the help of Tim Salimans, Alec Radford, Rif A. Saurous and others who have given us valuable feedback at various stages of writing.
\end{acknowledgements}

\appendix

\chapter{Appendix}
\label{chap:appendix}

\section{Notation and definitions}\label{sec:notdef}

\subsection{Notation}
\bgroup
\def\arraystretch{1.5}%
\begin{longtable}{| p{2.5cm} | p{8cm} |}
	\hline 
	\textbf{Example(s)} & \textbf{Description} \\
	\hline \hline
	$\bx,\by\,\bz$
	&
	With characters in bold we typically denote random \emph{vectors}. We also use this notation for collections of random variables variables.
	\\ \hline
	$x,y,z$
	&
	With characters in italic we typically denote random \emph{scalars}, i.e. single real-valued numbers.
	\\ \hline
	$\bX,\bY,\bZ$
	&
	With bold and capitalized letters we typically denote random \emph{matrices}.
	\\ \hline
	$Pa(\bz)$
	&
	The parents of random variable $\bz$ in a directed graph.
	\\ \hline
	$\text{diag}(\bx)$
	&
	Diagonal matrix, with the values of vector $\bx$ on the diagonal.
	\\ \hline
	$\bx \odot \by$
	&
	Element-wise multiplication of two vectors. The resulting vector is $(x_1\allowbreak y_1, ..., x_K y_K)^T$.
	\\ \hline
	$\theta$
	&
	Parameters of a (generative) model are typically denoted with the Greek lowercase letter $\theta$ (theta).
	\\ \hline
	$\bphi$
	&
	Variational parameters are typically denoted with the bold Greek letter $\bphi$ (phi).
	\\ \hline
	$p(\bx), p(\bz)$
	&
	Probability density functions (PDFs) and probability mass functions (PMFs), also simply called \emph{distributions}, are denoted by $p(.)$, $q(.)$ or $r(.)$.
	\\ \hline
	$p(\bx,\by,\bz)$
	&
	Joint distributions are denoted by $p(.,.)$
	\\ \hline
	$p(\bx|\bz)$
	&
	Conditional distributions are denoted by $p(.|.)$
	\\ \hline
	$p(.;\theta), p_\theta(\bx)$
	&
	The parameters of a distribution are denoted with $p(.;\theta)$ or equivalently with subscript $p_\theta(.)$.
	\\ \hline
	$p(\bx = \ba)$, $p(\bx \leq \ba)$
	&
	We may use an (in-)equality sign within a probability distribution to distinguish between function arguments and value at which to evaluate. So $p(\bx = \ba)$ denotes a PDF or PMF over variable $\bx$ evaluated at the value of variable $\ba$. Likewise, $p(\bx \leq \ba)$ denotes a CDF evaluated at the value of $\ba$.
	\\ \hline
	$p(.), q(.)$
	&
	We use different letters to refer to different probabilistic models, such as $p(.)$ or $q(.)$. Conversely, we use the \emph{same} letter across different marginals/conditionals to indicate they relate to the same probabilistic model.
	\\ \hline
\end{longtable}
\egroup

\subsection{Definitions}
\bgroup
\def\arraystretch{1.5}%
\begin{longtable}{| p{2.5cm} | p{8cm} |}
	\hline 
	\textbf{Term} & \textbf{Description}\\
	\hline \hline
	Probability density function (PDF) & A function that assigns a probability \emph{density} to each possible value of given \emph{continuous} random variables.
	\\ \hline
	Cumulative distribution function (CDF) & A function that assigns a cumulative probability density to each possible value of given univariate \emph{continuous} random variables.
	\\ \hline
	Probability mass function (PMF) & A function that assigns a probability \emph{mass} to given \emph{discrete} random variable.
	\\ \hline
\end{longtable}
\egroup

\subsection{Distributions}

We overload the notation of distributions (e.g. $p(\bx) = \mathcal{N}(\bx; \bmu, \boldsymbol{\Sigma})$) with two meanings: (1) a distribution from which we can sample, and (2) the probability density function (PDF) of that distribution.

\bgroup
\def\arraystretch{1.5}%
\begin{longtable}{| p{3cm} | p{7.5cm} |}
	\hline 
	\textbf{Term} & \textbf{Description}\\
	\hline \hline
	$\text{Categorical}(x; \bp)$ 
	& Categorical distribution, with parameter $\bp$ such that $\sum_i p_i = 1$.
	\\ \hline
	$\text{Bernoulli}(\bx; \bp)$ 
	& Multivariate distribution of independent Bernoulli.\\ & $\text{Bernoulli}(\bx; \bp) = \prod_i \text{Bernoulli}(x_i; p_i)$ with $\forall i: 0 \leq p_i \leq 1$.
	\\ \hline
	$\text{Normal}(\bx; \bmu, \boldsymbol{\Sigma}) = \mathcal{N}(\bx; \bmu, \boldsymbol{\Sigma})$ & Multivariate Normal distribution with mean $\bmu$ and covariance $\boldsymbol{\Sigma}$.
	\\ \hline
\end{longtable}
\egroup

\subsubsection{Chain rule of probability}\label{sec:chainruleprob}
\begin{align}
p(\ba,\bbb) = p(\ba)p(\bbb|\ba)
\end{align}

\subsubsection{Bayes' Rule}\label{sec:bayesrule}
\begin{align}
p(\ba|\bbb) = p(\bbb|\ba)p(\ba)/p(\bbb)
\end{align}

\subsection{Bayesian Inference}
\label{sec:bayesianinference}

Let $p(\theta)$ be a chosen marginal distribution over its parameters $\theta$, called a \emph{prior distribution}. Let $\mathcal{D}$ be observed data, $p(\mathcal{D}|\theta) \equiv \pT(\mathcal{D})$ be the probability assigned to the data under the model with parameters $\theta$. Recall the chain rule in probability:
\begin{align*}
p(\theta,\mathcal{D}) = p(\theta|\mathcal{D})p(\mathcal{D}) = p(\theta)p(\mathcal{D}|\theta)
\end{align*}
Simply re-arranging terms above, the posterior distribution over the parameters $\theta$, taking into account the data $\mathcal{D}$, is:
\begin{align}
p(\theta|\mathcal{D}) = \frac{p(\mathcal{D}|\theta)p(\theta)}{p(\mathcal{D})} \propto p(\mathcal{D}|\theta)p(\theta)
\label{eq:bayesrule}
\end{align}
where the proportionality ($\propto$) holds since $p(\mathcal{D})$ is a constant that is not dependent on parameters $\theta$. The formula above is known as \emph{Bayes' rule}, a fundamental formula in machine learning and statistics, and is of special importance to this work.

A principal application of Bayes' rule is that it allows us to make predictions about future data $\bx'$, that are optimal as long as the prior $p(\theta)$ and model class $\pT(\bx)$ are correct:
\begin{align*}
p(\bx=\bx'|\mathcal{D}) = \int \pT(\bx=\bx') p(\theta | \mathcal{D}) d \theta 
\end{align*}

\section{Alternative methods for learning in DLVMs}
\label{sec:altneratives}

\subsection{Maximum A Posteriori}
\label{sec:map}
From a Bayesian perspective, we can improve upon the maximum likelihood objective through \emph{maximum a posteriori} (MAP) estimation, which maximizes the log-posterior w.r.t. $\theta$. With i.i.d. data $\mathcal{D}$, this is:
\begin{align}
L^{MAP}(\theta)
&= \log p(\theta|\mathcal{D})\\
&= \log p(\theta) + L^{ML}(\theta)+ \text{constant}
\label{eq:map}\end{align}
The prior $p(\theta)$ in equation \eqref{eq:map} has diminishing effect for increasingly large $N$. For this reason, in case of optimization with large datasets, we often choose to simply use the maximum likelihood criterion by omitting the prior from the objective, which is numerically equivalent to setting $p(\theta) = \text{constant}$.

\subsection{Variational EM with local variational parameters}

Expectation Maximization (EM) is a general strategy for learning parameters in partially observed models~\citep{dempster1977em}. See section~\ref{sec:mcmcem} for a discussion of EM using MCMC. The method can be explained as coordinate ascent on the ELBO~\citep{neal1998em}. In case of of i.i.d. data, traditional variational EM methods estimate \textbf{local variational parameters} $\bphi^{(i)}$, i.e. a separate set of variational parameters per datapoint $i$ in the dataset. In contrast, VAEs employ a strategy with \textbf{global variational parameters}.

EM starts out with some (random) initial choice of $\bT$ and $\bphi^{(1:N)}$. It then iteratively applies updates:
\begin{align}
&\forall i = 1, ..., N:\;\; \bphi^{(i)} \leftarrow \argmax_{\bphi} \mathcal{L}(\bx^{(i)}; \bT,\bphi) &\text{\;\;(E-step)}\\
&\bT \leftarrow \argmax_{\bT} \sum_{i=1}^N \mathcal{L}(\bx^{(i)}; \bT,\bphi) &\text{\;\;(M-step)} 
\end{align}
until convergence. Why does this work? Note that at the E-step:
\begin{align}
&\argmax_{\bphi} \mathcal{L}(\bx; \bT,\bphi) \\
&= \argmax_{\bphi} \left[ \log \pT(\bx) - D_{KL}(\qP(\bz|\bx)||\pT(\bz|\bx)) \right]\\
&= \argmin_{\bphi} D_{KL}(\qP(\bz|\bx)||\pT(\bz|\bx))
\end{align}
so the $E$-step, sensibly, minimizes the KL divergence of $\qP(\bz|\bx)$ from the true posterior.

Secondly, note that if $\qP(\bz|\bx)$ equals $\pT(\bz|\bx)$, the ELBO equals the marginal likelihood, but that for any choice of $\qP(\bz|\bx)$, the $M$-step optimizes a bound on the marginal likelihood. The tightness of this bound is defined by $D_{KL}(\qP(\bz|\bx)||\pT(\bz|\bx))$.

\subsection{MCMC-EM}
\label{sec:mcmcem}
Another Bayesian approach towards optimizing the likelihood $\pT(\bx)$ with DLVMs is Expectation Maximization (EM) with Markov Chain Monte Carlo (MCMC). In case of MCMC, the posterior is approximated by a mixture of a set of approximately i.i.d. samples from the posterior, acquired by running a Markov chain. Note that posterior gradients in DLVMs are relatively affordable to compute by differentiating the log-joint distribution w.r.t. $\bz$:
\begin{align}
\nabla_\bz \log \pT(\bz | \bx) 
&= \nabla_\bz \log [\pT(\bx, \bz)/\pT(\bx)]\\
&= \nabla_\bz [\log \pT(\bx, \bz) - \log \pT(\bx)]\\
&= \nabla_\bz \log \pT(\bx, \bz) - \nabla_\bz \log \pT(\bx)\\
&= \nabla_\bz \log \pT(\bx, \bz)
\end{align} 
One version of MCMC which uses such posterior for relatively fast convergence, is Hamiltonian MCMC~\citep{neal2011mcmc}. A disadvantage of this approach is the requirement for running an independent MCMC chain per datapoint.

\section{Stochastic Gradient Descent}\label{sec:sgd}

We work with directed models where the objective per datapoint is scalar, and due to the differentiability of neural networks that compose them, the objective is differentiable w.r.t. its parameters $\theta$. 
Due to the remarkable efficiency of reverse-mode automatic differentiation (also known as the backpropagation algorithm~\citep{rumelhart1988learning}), the value and gradient (i.e. the vector of partial derivatives) of differentiable scalar objectives can be computed with equal time complexity. 
In SGD, we iteratively update parameters $\theta$:
\begin{align}
\theta_{t+1} \leftarrow \theta_t + \alpha_t \cdot \nabla_\theta \tilde{L}(\theta, \xi)
\label{eq:sgd}
\end{align}
where $\alpha_t$ is a learning rate or preconditioner, and $\tilde{L}(\theta,\xi)$ is an unbiased estimate of the objective $L(\theta)$, i.e. $\Exp{\xi \sim p(\xi)}{\tilde{L}(\theta, \xi)} = L(\theta)$. The random variable $\xi$ could e.g. be a datapoint index, uniformly sampled from $\{1,...,N\}$, but can also include different types of noise such posterior sampling noise in VAEs. In experiments, we have typically used the Adam and Adamax optimization methods for choosing $\alpha_t$~\citep{kingma2015adam}; these methods are invariant to constant rescaling of the objective, and invariant to constant re-scalings of the individual gradients. As a result, $\tilde{L}(\theta, \xi)$ only needs to be unbiased up to proportionality. We iteratively apply eq.~\eqref{eq:sgd} until a stopping criterion is met. A simple but effective criterion is to stop optimization as soon as the probability of a holdout set of data starts decreasing; this criterion is called \emph{early stopping}.

\backmatter

\printbibliography

\end{document}